\documentclass[runningheads]{llncs}

\usepackage{eccv}

\usepackage{eccvabbrv}
\usepackage{algorithmic}
\usepackage{algorithm}
\usepackage{graphicx}
\usepackage{booktabs}
\usepackage{multirow} 
\usepackage{multicol}
\usepackage{subcaption}
\usepackage{longtable}
\usepackage{booktabs}
 \newsavebox\CBox
\def\textBF#1{\sbox\CBox{#1}\resizebox{\wd\CBox}{\ht\CBox}{\textbf{#1}}}
\usepackage[accsupp]{axessibility}  % Improves PDF readability for those with disabilities.

\usepackage{hyperref}

\usepackage{orcidlink}

\begin{document}
 
\title{EntAugment: Entropy-Driven Adaptive Data Augmentation Framework for Image Classification} 

\titlerunning{EntAugment: Entropy-Driven Adaptive Data Augmentation Framework}

\author{Suorong Yang\inst{1,2}\orcidlink{0000-0001-8788-6382} \and
Furao Shen\inst{1,3}\orcidlink{0000-0002-7285-326X}\dag
\and
Jian Zhao\inst{4}\orcidlink{0000-0003-3949-352X}}
\footnotetext[2]{$^\dag$ Corresponding authors.}
\authorrunning{S.~Yang et al.}

\institute{State Key Laboratory for Novel Software Technology, Nanjing University, China\and
Department of Computer Science and Technology, Nanjing University, China \and
School of Artificial Intelligence, Nanjing University, China \and
School of Electronic Science and Engineering, Nanjing University, China\\
\email{sryang@smail.nju.edu.cn}, 
\email{\{frshen, jianzhao\}@nju.edu.cn}}

\maketitle

\begin{abstract}
 Data augmentation (DA) has been widely used to improve the generalization of deep neural networks.
 While existing DA methods have proven effective, they often rely on augmentation operations with random magnitudes to each sample.
 However, this approach can inadvertently introduce noise, induce distribution shifts, and increase the risk of overfitting.
In this paper, we propose EntAugment, a tuning-free and adaptive DA framework.
Unlike previous work, EntAugment dynamically assesses and adjusts the augmentation magnitudes for each sample during training, leveraging insights into both the inherent complexities of training samples and the evolving status of deep models.
Specifically, in EntAugment, the magnitudes are determined by the information entropy derived from the probability distribution obtained by applying the softmax function to the model's output.
In addition, to further enhance the efficacy of EntAugment, we introduce a novel entropy regularization term, EntLoss, which complements the EntAugment approach.
Theoretical analysis further demonstrates that EntLoss, compared to traditional cross-entropy loss, achieves closer alignment between the model distributions and underlying dataset distributions.
Moreover, EntAugment and EntLoss can be utilized separately or jointly.
We conduct extensive experiments across multiple image classification tasks and network architectures with thorough comparisons of existing DA methods.
Importantly, the proposed methods outperform others without introducing any auxiliary models or noticeable extra computational costs, highlighting both effectiveness and efficiency.
Code is available at \url{https://github.com/Jackbrocp/EntAugment}.
\end{abstract}

\section{Introduction}
\label{sec:intro}
Data augmentation (DA) has been widely utilized in training deep neural networks to alleviate overfitting and enhance models' generalization performance~\cite{survey2, survey, survey3,naveed2024survey,DA_snn}.
% However, we revisit the design of the prevalent DA framework and argue that it is a suboptimal strategy for model training.
However, the prevalent DA frameworks employed for training deep networks commonly adhere to a statically invariant strategy, which is uniformly applied to all training samples throughout the training process~\cite{cutout,autoaugment,trivialaugment, randaugment, fast-autoaugment,advmask,dada}.
Regardless of specific augmentation operations employed, these methods typically incorporate randomly sampled parameters to adjust the strengths of augmentation.
Consequently, the variability in the augmented data is stochastic and non-adaptive to training samples or target model training status.
For instance, information deletion-based DA methods~\cite{randomerasing,has,advmask,gridmask} adopt deterministic strategies to randomly erase some information in images during training.
Moreover, automatic DA approaches~\cite{autoaugment,fast-autoaugment,randaugment,trivialaugment,selectaugment} search for the augmentation operation space and parameter space prior to the actual training.
During online training, these methods utilize randomly chosen operations and parameters to generate augmented data, thus non-adaptive to different samples and model training progression.
Meanwhile, for different datasets and models, the augmentation parameters have to be manually customized or searched before task model training~\cite{randaugment,autoaugment,cutout,advmask,deepaa}, which hinders practicality.

Despite the efficacy, existing DA methods often neglect the critical aspect of customizing augmentation magnitudes for individual images based on the model training progression, leading to uncontrolled variations in training data.
More importantly, the uncontrolled variations in training data can potentially lead to severe drawbacks~\cite{investigating}.
Specifically, such variations in the current DA frameworks may inevitably result in excessive or insufficient data manipulation, as evidenced in previous studies~\cite{advmask,keepaugment,selectaugment}.
Excessive data manipulation engenders significant variations, introducing noisy samples and distribution shifts.
Conversely, insufficient data manipulation may elevate the overfitting risks.
Thus, if the augmentation is not aptly modulated, it can deteriorate the overall model performance, which underscores the intrinsic limitations of the existing DA mechanisms.

\begin{figure}[t]
	\centering
         \includegraphics[width=7cm]{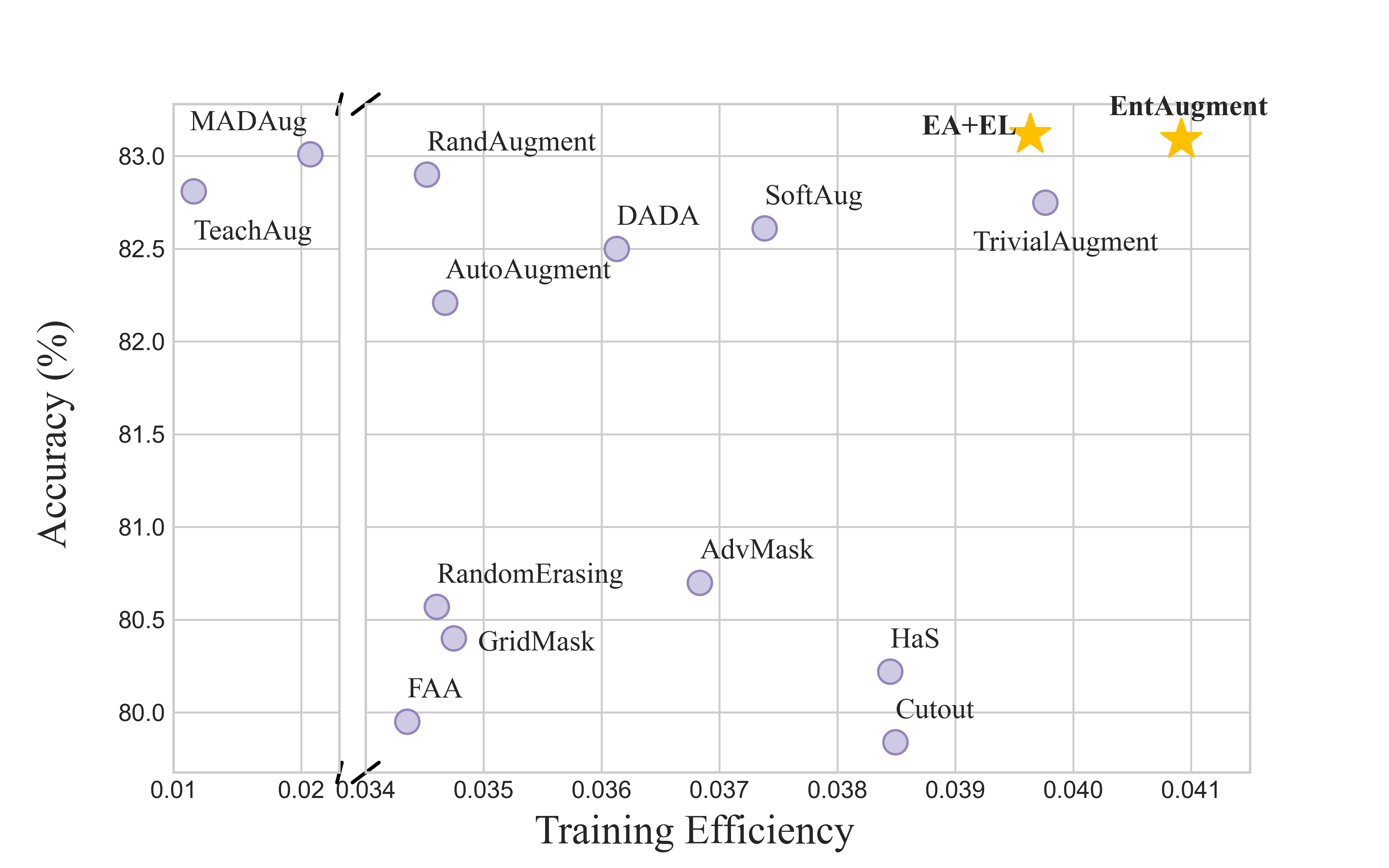}
	\caption{\textBF{Training efficiency \textit{vs.} classification accuracy.} Training efficiency is calculated using the inverse of the average per epoch training time costs, where higher values are preferable. Regarding the efficiency-effectiveness trade-off, our method performs better than prior SOTA DA methods. EA+EL: EntAugment+EntLoss.}
    \label{fig:intro}
\end{figure}
% 这里有一个问题，没有提到training status的问题
% 方法的设计和训练状态无关
Given these challenges, it is beneficial to dynamically customize the extent of DA manipulation for each sample.
This customization should take into account both the sample's inherent difficulty and the models' current training status (i.e., generalization capacity).
For instance, in the early phases of training, when the model's capacity is limited, and it struggles to classify the majority of samples correctly, augmenting training data with slight magnitudes can expedite performance improvement.
As the model progresses and enhances its ability to generalize, a notable portion of samples becomes easier to classify.
In this stage, intensifying the augmentation manipulations helps provide augmented data in more diverse scenarios, thereby enabling models to capture more discriminative features.
This adaptive DA framework also aligns with the principles of curriculum learning~\cite{curriculum,curriculum2,curriculum3,curriculum4} and has demonstrated superior efficacy in the context of model training~\cite{advmask,gridmask,hou2023learn}.

In this paper, we propose EntAugment, a novel, tuning-free, and adaptive DA framework.
EntAugment incorporates a dynamic adjustment mechanism based on the inherent complexities of individual training samples and the evolving state of models throughout training.
However, an accurate and efficient assessment of sample difficulty during online training poses a challenge, primarily due to the limitations in dynamically assessing the model's training progression based solely on data-derived difficulty measures.
To address this challenge, we introduce an entropy-driven data augmentation mechanism that leverages information entropy derived from the probability distribution obtained by applying the softmax function to the model's output.
Although straightforward in concept, the entropy metric derived for each sample dynamically evolves throughout model training.
It serves as a robust indicator of the model's classification confidence for individual samples.
More specifically, samples with inherent complexity and those encountered early in training typically exhibit higher entropy values, indicating greater difficulty.
As training advances, samples that are initially challenging may exhibit reduced difficulty over time, highlighting dynamic shifts in difficulty measures.
Consequently, determining sample difficulty becomes an adaptive and dynamic process shaped by the interplay between the model's evolving capabilities and the training data's inherent characteristics.
In EntAugment, more challenging samples receive more conservative augmentations to enhance the model's learning while mitigating the introduction of excessive noise.
Conversely, simpler samples benefit from more extensive variations, thereby reducing the risk of overfitting. 
Meanwhile, it is also noteworthy that EntAugment only incurs minimal computational overhead and obviates the need for any auxiliary models or extra training overhead, highlighting its superior efficiency.
 
To further enhance the effectiveness of EntAugment, we extend EntAugment by introducing an entropy regularization term, denoted as EntLoss, into the standard cross-entropy (CE) loss. 
EntLoss is primarily engineered to bolster classification confidence (i.e., minimizing entropy) and enhance the overall performance of EntAugment.
Furthermore, EntLoss is underpinned by a rigorous theoretical analysis, wherein we prove that it helps models better fit the underlying dataset distribution (see Proposition~\ref{prop1}).
Precisely, the dissimilarity induced by the vanilla CE loss serves as an upper bound for the dissimilarity induced by the combined application of CE loss and EntLoss.
Therefore, EntLoss brings additional benefits of enhancing the model's ability to fit the dataset distribution, thereby improving generalization performance and surpassing the efficacy of existing approaches based on vanilla CE loss.
Experiments across a variety of deep models and datasets, e.g., CIFAR-10\slash 100~\cite{cifar}, Tiny-ImageNet~\cite{tiny}, ImageNet~\cite{imagenet}, ImageNet-LT~\cite{lt-datasets}, Places-LT~\cite{lt-datasets} demonstrate that our methods bring greater improvement to task models than existing SOTA DA methods in terms of test-set performance while incurring minimal training cost.

% achieve superior performance compared to other SOTA DA methods while incurring only a negligible increase in computational overhead, highlighting efficiency.

% Specifically, EntLoss is underpinned by a rigorous theoretical analysis, wherein we prove that the incorporation of EntLoss helps deep models fit the dataset distribution more accurately.
% Notably, the dissimilarity induced by the vanilla CE loss serves as an upper bound for the dissimilarity introduced by CE loss and EntLoss. 
% Empirically, EntLoss demonstrates substantial improvements in the performance of image classification models, surpassing the performance of existing approaches based on CE loss.
 
Our main contributions can be summarized as follows:
\begin{enumerate}
    \item We propose EntAugment, a tuning-free and adaptive DA framework, which dynamically adjusts the magnitudes of DA for individual training samples based on sample difficulty and the model training progression.
    \item To the best of our knowledge, we are the first to pinpoint that the sample difficulty is not static but varies through model training.
    This is used to mitigate the side effects caused by the uncontrollable randomness of DA, which is overlooked in prior works.
    \item We introduce EntLoss, an entropy regularization loss term for classification model training, accompanied by theoretical foundations that elucidate its role in enhancing model alignment with dataset distribution, thereby contributing to generalization capabilities. 
    \item EntAugment and EntLoss can be employed both separately and jointly. Extensive experimental results demonstrate the superior effectiveness and efficiency of our methods compared to existing SOTA DA methods.
\end{enumerate}

\section{Related Work}
\label{sec:related}
\subsection{Data Augmentation}
Recently, many data augmentation methods have emerged with the primary objective of enhancing the generalization capabilities of deep neural networks.
% These methods typically employ a fixed strategy that remains consistent throughout the training process.
Image erasing-based data augmentation approaches typically erase some random information within images.
For instance, Cutout~\cite{cutout}, GridMask~\cite{gridmask}, HaS~\cite{has}, and Random Erasing~\cite{randomerasing} randomly mask out one or more structural regions within images.
% Cutout~\cite{cutout} randomly masks out one or more square regions within images.
%  GridMask~\cite{gridmask} focuses on erasing structural regions within images, thereby mitigating the continuous and excessive removal of information. 
% Similarly, Random Erasing~\cite{randomerasing} randomly selects a rectangular region in an image and replaces its pixels with random values, thereby mitigating the risk of overfitting.
Since these methods neglect the structural characteristics of training images, AdvMask~\cite{advmask} identifies the critical regions within images offline and selectively drops some structural sub-regions with critical points during online augmentation.
However, these methods neglect the training dynamics of deep models.
This oversight may inadvertently hamper the optimization potential and efficacy of the learning process, resulting in suboptimal performance.

Image mixing-based data augmentation methods, such as Mixup~\cite{mixup} and CutMix~\cite{cutmix}, typically mix random information from two or more images to generate the augmented data during training.
While effective in generating diverse augmented data, the magnitude of variations introduced by these methods is random, which may inevitably introduce distribution shifts and noises.
Meanwhile, the above methods need expert knowledge to design the operations and parameters for specific datasets~\cite{dada}.

Automated data augmentation methods have demonstrated superior performance~\cite{autoaugment,selectaugment,teachaugment,adversarial-aa,trivialaugment,dada}.
These methods typically try to search augmentation policies and parameters automatically based on some metrics before task model training.
For instance, AutoAugment~\cite{autoaugment} employs reinforcement learning to search for the optimal combination of data augmentation policies tailored to individual datasets. 
 Fast-AutoAugment~\cite{fast-autoaugment}, motivated by density matching principles between training and test datasets, introduces an inference-only metric for the evaluation of data augmentation operations.
RandAugment~\cite{randaugment} leverages grid search to select a combination of augmentation operations for various datasets.
AWS~\cite{improveautoaugment} designs an augmentation-wise weight-sharing strategy to search augmentation operations.
In contrast, TrivialAugment~\cite{trivialaugment} adopts the same augmentation space obtained by these methods but opts for a simpler approach by applying a single augmentation operation to each image during training.
SelectAugment~\cite{selectaugment} utilizes a two-step Markov decision process and hierarchical reinforcement learning to learn the augmentation policy and select samples to be augmented online. 
However, as SelectAugment employs AutoAugment, Mixup, or CutMix for data augmentation, the strengths of the augmentation applied remain uncontrollable.
Moreover, AdDA~\cite{adda} is proposed for contrastive learning, which learns to adaptively adjust the augmentation compositions and achieves more generalizable representations.
SoftAug~\cite{softaug} generates augmentation with invariant transforms to soft augmentations.
MADAug~\cite{hou2023learn} jointly trains an augmentation policy network through a bi-level optimization scheme to select augmentation operations for samples.
TeachAugment~\cite{teachaugment} leverages a teacher model to generate the transformed data based on the adversarial strategy.
DADA~\cite{dada} relaxes the discrete DA policy selection to a differentiable optimization problem, facilitating efficient and accurate DA policy learning.
Since these methods adopt augmentation operations with random or fixed parameters, the magnitude of variations in the augmented data is difficult to adjust.
Our proposed method distinguishes itself by adaptively determining the augmentation magnitudes based on image characteristics and the dynamics of the model training process, thereby enhancing the flexibility and efficacy of data augmentation.
% Consequently, our method empowers the adaptive determination of augmentation magnitudes for each image, thereby enhancing the flexibility and efficacy of the data augmentation.

\subsection{Entropy-Regularized Loss}
In the realm of deep learning and machine learning,  entropy-regularized loss functions are typically designed to encourage models to produce predictions that are more evenly spread or uncertain~\cite{loss1,loss2}.
Particularly within the domains of reinforcement learning and decision-making processes, entropy regularization finds widespread application to improve the policy optimization process, e.g., obtaining high-entropy output to encourage exploration~\cite{RLloss1,RLloss2,RLloss3}.
Minimum entropy regularizers have been used in other contexts, such as semi-supervised learning problem~\cite{ER1}, unsupervised clustering~\cite{ER2}, structured output prediction~\cite{ER3}, and unsupervised domain adaption~\cite{entropy-mini}, etc.
In contrast, our work advances the proposition of integrating entropy-regularized loss into the conventional CE loss to expedite model convergence and mitigate model fitting errors.

\section{The Proposed Method}
In this section, we first propose EntAugment. We then introduce the entropy-regularized loss EntLoss and study how it improves the performance of EntAugment and the overall generalization capabilities of task models.
\subsection{EntAugment}
EntAugment is motivated by a straightforward intuition that conventional data augmentation, which employs a uniform strategy characterized by random magnitudes, may lead to sub-optimal performance.
Regardless of specific DA operations, it is imperative to dynamically adjust the magnitude of these operations in accordance with the progress of the target model training and the inherent difficulty of individual samples.
For instance, for samples that are classified by the current target model with high confidence, suggesting relative ease of classification, augmentation should encompass a broad spectrum of scenarios to diversify the training data.
Conversely, samples that models struggle to classify with high confidence scores, indicating a significant challenge in learning, necessitate more subtle variations in augmentation. 
Nevertheless, it is essential to acknowledge that the effective and efficient assessment of sample difficulty, while concurrently considering the model's training status, poses a formidable challenge.
In light of this challenge, we propose EntAugment.
% This strategy is designed to facilitate the model's in-depth learning of the intricate features of such challenging samples.
% To this end, we propose an entropy-driven adaptive data augmentation method.
% \begin{algorithm}[t]
% \caption{General EntAugment Procedure}
% \label{alg}
% \begin{algorithmic}[1]
% \REQUIRE {a batch of image and class label pairs $(\boldsymbol{x},y)$, batch size $bs$, total number of classes $k$, a classification model at time $t$, $f_t$, an augmentation space $\mathcal{A}$}
% \ENSURE a batch of augmented image $\boldsymbol{x}'$
% \FOR{$i=0$:$bs-1$}
%     \STATE Calculate the magnitude of the DA manipulation for $\boldsymbol{x}_i$ based on Eq.~\eqref{eq:mag}, denoted as $mag_i$
%     \STATE Randomly sample an augmentation operation $\boldsymbol{a}$ from $\mathcal{A}$
%     \STATE Augment sample $\boldsymbol{x'}_i$:  $\boldsymbol{x'}_i = \boldsymbol{a}(\boldsymbol{x}_i, mag_i)$
% \ENDFOR
% % \STATE $\boldsymbol{m}_T$ $\leftarrow$ sigmoid($\alpha_T \cdot \mathcal{H}_T(\boldsymbol{\delta}_T)$)
% % \STATE $\boldsymbol{\delta}$ $\leftarrow$ $\boldsymbol{\delta_T}\odot \boldsymbol{m}_T$
% \RETURN a batch of augmented data $\boldsymbol{x'}$
% \end{algorithmic}
% \end{algorithm}

EntAugment leverages the augmentation space that has been utilized in previous works~\cite{autoaugment,trivialaugment,selectaugment}.
Let $\mathcal{D}$ denote the training dataset comprising $N$ training samples, each of the form $(\boldsymbol{x},y) \in \mathcal{D}$.
$\boldsymbol{x}$ represents the original data, and $y=1,...,k$ is the corresponding label, where $k$ is the total number of classes.
Given a classification model $f_{\theta}$ parameterized by $\theta$ and an input sample $\boldsymbol{x} \in \mathbb{R}^n$, $f_{\theta}(\boldsymbol{x})$ is the network output.
For simplicity, we use $g(\boldsymbol{x}) = \mathrm{softmax}(f_{\theta}(\boldsymbol{x})) \in \mathbb{R}^k$ to denote the output of the softmax function.
Therefore, $g(\boldsymbol{x}) $ is indeed a probability distribution, i.e., $\sum_{i=1}^k g(\boldsymbol{x})_i = 1$.
The information entropy of the softmax output is defined as:
\begin{equation}\label{eq:entropy}
    -\sum_{i=1}^k g(\boldsymbol{x})_i \log g(\boldsymbol{x})_i,
\end{equation}
which indicates the confidence level of model $f_{\theta}$ in classifying $\boldsymbol{x}$.
In cases where samples pose challenges for classification, the entropy associated with the model's output tends to exhibit relatively higher values, whereas lower values indicate easily classifiable data.
One notable advantage of the entropy measure defined in Eq.~\eqref{eq:entropy} lies in its \textBF{dual functionality} - encapsulating information regarding sample complexity and insights into model evolution during training. 
Therefore, dynamic adjustment of DA magnitude becomes viable.
Specifically, the magnitude of the DA manipulation is determined by:
\begin{equation}\label{eq:mag}
mag(\boldsymbol{x}) =    1 + \frac{1}{\log k} \sum_{i=1}^k g(\boldsymbol{x})_i \log g(\boldsymbol{x})_i .
\end{equation}
In this way, $mag(\boldsymbol{x})$ scales to $[0,1]$ and exhibits an inverse proportional relationship with the entropy measures. 
In situations where $mag(\boldsymbol{x}) \rightarrow 1$, the augmented samples exhibit a greater degree of variability, while conversely, minor alterations occur as $mag(\boldsymbol{x}) \rightarrow 0$.
Additionally, it is noteworthy to highlight that $mag(\boldsymbol{x})$ for each sample $\boldsymbol{x}$ remains dynamic and evolves continuously throughout the entire training process, \textBF{enabling adaptive augmentation}. 

\begin{figure}[t]
    \centering
    \begin{minipage}{0.6\textwidth}
     \begin{algorithm}[H]
    \caption{General EntAugment Procedure}\label{alg:ent}
    \label{alg}
    \begin{algorithmic}
    \REQUIRE {a batch of image and class label pairs $(\boldsymbol{x},y)$, batch size $bs$, total number of classes $k$, an augmentation space $\mathcal{A}$}
    \ENSURE a batch of augmented image $\boldsymbol{x}'$
    \FOR{$i=0$: $bs-1$}
        \STATE Calculate the magnitude of the DA manipulation for $\boldsymbol{x}_i$ based on Eq.~\eqref{eq:mag}, denoted as $mag_i$
        \STATE Randomly sample an augmentation operation $\boldsymbol{a}$ from $\mathcal{A}$
        \STATE Augment sample $\boldsymbol{x}_i$:  $\boldsymbol{x'}_i = \boldsymbol{a}(\boldsymbol{x}_i, mag_i)$
    \ENDFOR
    \RETURN a batch of augmented data $\boldsymbol{x'}$
    \end{algorithmic}
    \end{algorithm}
    \end{minipage}\hfill
    \begin{minipage}{0.38\textwidth}
        \centering
         \includegraphics[width=4.5cm]{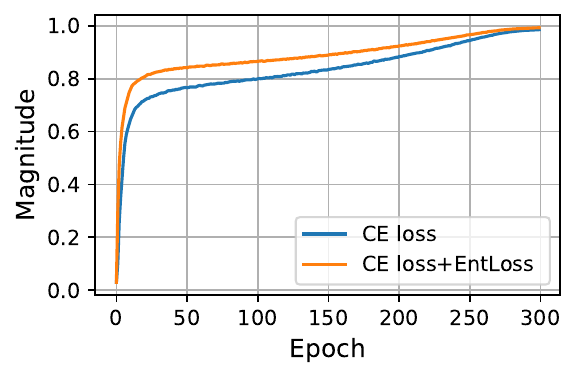}
	\caption{Values of magnitude defined in Eq.~\eqref{eq:mag} with and without employing EntLoss over the training process.}
    \label{fig:entloss}
    \end{minipage}
\end{figure}

Based on Eq.~\eqref{eq:mag}, during the initial phase of model training, most samples are difficult to classify, thereby assigned with relatively low magnitudes.
Providing relatively simple samples during the prior stage of model training contributes to the enhancement of overall model performance~\cite{curriculum,curriculum2}.
As models acquire enhanced generalization capabilities, the entropy values in Eq.~\eqref{eq:entropy} tend to decrease for more data, indicating higher magnitudes defined by Eq.~\eqref{eq:mag}. 
More augmented data will be provided in more diverse scenarios.
Thus, the adaptive data augmentation framework can be ensured.
The procedure of EntAugment is outlined in Algorithm~\ref{alg:ent}.
% \begin{figure}[]
% 	\centering
%          \includegraphics[width=5cm]{./pictures/entloss.pdf}
% 	\caption{Values of magnitude defined in Eq.~\eqref{eq:mag} with and without employing EntLoss over the training process.}
%     \label{fig:entloss}
% \end{figure}

\subsection{Entropy-Regularized Loss}\label{sec:entloss}
Although EntAugment performs well theoretically,
% Despite EntAugment proposing an entropy-driven adaptive DA framework,
the model's confidence scores in data classification may slowly increase and stabilize at a moderate level.
As illustrated in Figure~\ref{fig:entloss}, the magnitudes defined in Eq.~\eqref{eq:mag} and derived using vanilla CE loss remain consistently low throughout most training epochs.
Nonetheless, this could lead to inadequate DA magnitudes and a lack of potential diversity in the augmented data, particularly during later stages of model training.

To address this issue, based on Eq.~\eqref{eq:entropy} and Eq.~\eqref{eq:mag}, we introduce an entropy regularization loss, which is defined as:
\begin{equation}\label{eq:entloss}
    \mathrm{EntLoss}(\boldsymbol{x}) = \frac{1}{\log k} \sum_{i=1}^k g(\boldsymbol{x})_i \log g(\boldsymbol{x})_i.
\end{equation}
When incorporated with vanilla CE loss, EntLoss encourages models to classify samples with higher confidence, i.e., the models trained with EntLoss attain lower entropy values.
Let $\hat{\theta}_{cro}$ denote the empirical risk minimizer employing the conventional CE loss, and $\hat{\theta}_{ent}$ be the empirical risk minimizer utilizing both the CE loss and EntLoss.
Given that the entropy function is denoted as $\mathrm{Ent}(\cdot)$, we have
% Compared with vanilla cross-entropy loss, the models trained with entropy-regularized loss approximate distribution with lower entropy:
\begin{equation}\label{eq:ent}
    \mathrm{Ent}(p_{\hat{\theta}_{cro}}) \leq \mathrm{Ent}(p_{\hat{\theta}_{ent}}),
\end{equation}
where $p_{\theta}$ is the probability distribution of the model $f_{\theta}$'s output. 
Furthermore, as shown in Figure~\ref{fig:entloss}, the confidence level and magnitudes increase substantially after incorporating EntLoss into the training process.
% In Figure~\ref{fig:entloss}, after incorporating EntLoss into the training process, the confidence level and magnitude increase substantially, which ensures the diversity of training data.
More importantly, EntLoss brings more significant advantages in training classification models.
\subsubsection{Theoretical Analysis}
We will theoretically demonstrate that EntLoss can mitigate dissimilarity between the distributions of the model and dataset compared to the conventional CE loss.
This facilitates the model to fit the dataset distribution better, thereby enhancing its generalization capability.

Our initial exposition delves into the equivalence between the effect of the vanilla CE loss function and the minimization of the Kullback-Leibler (KL) divergence between the model's and the dataset's distributions.
% We first show that the effect of the vanilla cross-entropy loss function is equivalent to minimizing the Kullback-Leibler (KL) divergence between the model and dataset distribution.
Consider a dataset denoted as $\mathcal{D}=\{\mathbf{z}_1, ..., \mathbf{z}_N\}$, comprising $N$ training instances, where each $\mathbf{z}_i=(\boldsymbol{x}_i, y_i )$ is independently generated from an unknown potential data distribution, denoted as $p_{data}(\mathbf{z})$.
Further, let $p_{model}(\boldsymbol{\theta})$ represent a parametric family of probability distributions over the same space indexed by $\boldsymbol{\theta}$.
We have the following Lemma.
% containing $N$ training samples where $\mathbf{z}_i=(\boldsymbol{x}_i, y_i )$ is independently generated from an unknown true data distribution $p_{data}(\mathbf{z})$.
% Let $p_{model}(\boldsymbol{\theta})$ be a parametric family of probability distributions over the same space indexed by $\boldsymbol{\theta}$. We have the following Lemma.
\begin{lemma}\label{lemma1}
    From the perspective of maximum likelihood estimation, the empirical risk minimizer can be denoted as $\hat{\boldsymbol{\theta}} =  \arg \max_{\boldsymbol{\theta}} \mathbb{E}_{\mathbf{z} \sim \hat{p}_{\text{data }}} \log p_{\text {model }}(\boldsymbol{z}; \boldsymbol{\theta})$, where $\hat{p}_{data}$ is the empirical distribution defined by the training set.
\end{lemma}
 Lemma~\ref{lemma1} is formally proven in Appendix B.
 Based on Lemma~\ref{lemma1}, the process of model training can be comprehended as minimizing the dissimilarity between the empirical distribution $\hat{p}_{data}$ and the model distribution $p_{model}$.
 Furthermore, the extent of dissimilarity between these two distributions can be effectively measured by the KL divergence, which is given by:
\begin{equation}\label{eq:KL}
    D_{\mathrm{KL}}\left(\hat{p}_{\text {data }} \| p_{\text {model }}\right)=\mathbb{E}_{\mathbf{z} \sim \hat{p}_{\text {data }}}\left[\log \hat{p}_{\text {data }}(\boldsymbol{z})-\log p_{\text {model }}(\boldsymbol{z})\right].
\end{equation}
$\mathbb{E}_{\mathbf{z} \sim \hat{p}_{\text {data }}}\left[\log \hat{p}_{\text {data }}(\boldsymbol{z})\right]$ is a function only of the training set, not the model.
When we train the model to minimize the KL divergence, we need to minimize 
\begin{equation}
    -\mathbb{E}_{\boldsymbol{z} \sim \hat{p}_{\text {data }}}\left[\log p_{\text {model }}(\boldsymbol{z})\right],
\end{equation}
which can be estimated by $-\frac{1}{n} \sum_{i=1}^n p_{\text {data}}(\boldsymbol{z}_i) \log p_{\text {model }}(\boldsymbol{z}_i)$.
Therefore, the following Lemma can be derived.
\begin{lemma}
    Minimizing the cross entropy loss is equivalent to minimizing the KL divergence between the model's distribution and the dataset's distribution, as shown in Eq.~\eqref{eq:KL}.
    % , as represented in Eq.~\eqref{eq:KL}, between the model's and the dataset's distribution.
\end{lemma}

However, the model's output is indeed a continuous variable, i.e., $\boldsymbol{\theta}(\boldsymbol{z}_i) \in \mathbb{R}^k, i=1,2,...,n$.
In contrast, the distribution of the dataset is structured as a one-hot variable.
For example, if $\boldsymbol{x}_i$ belongs to the $k'$-th class, $y_i = \left[0,...,1,...,0 \right]$, where $y_{ij}=1$ if $j=k'$ and  $y_{ij}=0$, otherwise.
This intrinsic inconsistency between the model's output and the dataset distribution gives rise to a notable theoretical discrepancy.
Thus, EntLoss can be used to complement the conventional CE loss function, making the model distribution closer to the data distribution. 
\begin{lemma}\label{lemma3}
Suppose that $\hat{\theta}_{cro}$ is the empirical risk minimizer using the vanilla CE loss and $\hat{\theta}_{ent}$ is the empirical risk minimizer utilizing the CE loss along with the EntLoss, we have
\begin{equation}\label{eq:cross-ent-com}
    -\mathbb{E}_{\mathbf{z} \sim \hat{p}_{\text {data }}}\left[\log p_{\hat{\theta}_{ent}}(\boldsymbol{z})\right] \leq -\mathbb{E}_{\mathbf{z} \sim \hat{p}_{\text {data }}}\left[\log p_{\hat{\theta}_{cro}}(\boldsymbol{z})\right]
\end{equation}
\end{lemma}
\begin{proposition}\label{prop1}
Based on Eq.~\eqref{eq:ent}, ~\eqref{eq:KL}, and Lemma~\ref{lemma3}, we have
\begin{equation}
    D_{KL}(\hat{p}_{data} || p_{\hat{\theta}_{cro}}) \geq D_{KL}(\hat{p}_{data} || p_{\hat{\theta}_{ent}}).
\end{equation}
\end{proposition}
Proposition~\ref{prop1} is proven in Appendix C.
It can be seen that the dissimilarity between the model distribution $p_{\hat{\theta}_{cro}}$ and the underlying dataset distribution serves as an upper bound for the dissimilarity between the model distribution $p_{\hat{\theta}_{ent}}$ and the dataset distribution.
Proposition~\ref{prop1} reveals the advantageous effect of EntLoss in mitigating the dissimilarity between the model's distribution and the dataset's distribution.
Thus, EntLoss facilitates models to better fit the inherent distribution of the dataset, thereby achieving enhanced generalization.

In conclusion, EntLoss can be employed to enhance EntAugment or utilized as a standalone optimization method for training classification models.

\subsubsection{Complexity Analysis of EntLoss and EntAugment.}\label{sec:time_cost}
 We provide theoretical analysis demonstrating that the utilization of EntLoss and EntAugment does not introduce any notable additional computational cost.
 Specifically, the computational complexity of CE loss is $O(k \times N + N)$, where $N$ denotes the number of samples and $k$ is the number of categories. 
 Both EntAugment and EntLoss exhibit a time complexity of $O(k \times N)$.
 Thus, the computational complexity of EntAugment alone is closely equivalent to that of the vanilla CE loss, highlighting its efficiency.
 Meanwhile, the overall computational complexity is $O((2k+1)\times N)$, which is equivalent to the vanilla CE loss.

\section{Experiment}
% This section demonstrates the superior performance of EntAugment and EntLoss.
% \noindent \textBF{Experimental Settings}
% We evaluate the effectiveness and efficiency of our methods on six publicly available benchmark datasets. 
% First, we evaluate the performance of our methods on standard image classification datasets: CIFAR10/100~\cite{cifar}, Tiny-ImageNet~\cite{tiny}, and ImageNet~\cite{imagenet}.
% Besides standard image classification datasets, in the supplementary material, we further evaluate the generality of our methods on large-scale long-tailed datasets, including ImageNet-LT~\cite{lt-datasets} and Places-LT~\cite{lt-datasets}.
% Third, we employ transfer learning to compare the performance of various DA methods.
% Next, to demonstrate the effectiveness of our methods in enhancing model generalization, we visualize the deep features learned from CIFAR-10.
% We also provide the convergence analysis of our methods.
% In addition, we compare the training efficiency of various DA methods.
% Ablation studies are also conducted to show how the components affect the performance.

\subsubsection{Comparison with state-of-the-arts}
We compare our methods with the 11 most representative and commonly used data augmentation methods, including HaS~\cite{has}, Cutout~\cite{cutout}, CutMix~\cite{cutmix}, GridMask (GM)~\cite{gridmask}, AdvMask (AM)~\cite{advmask}, Random Erasing (RE)~\cite{randomerasing}, AutoAugment (AA)~\cite{autoaugment}, Fast-AutoAugment (FAA)~\cite{fast-autoaugment}, RandAugment (RA)~\cite{randaugment}, DADA~\cite{dada}, TrivialAugment (TA)~\cite{trivialaugment}, TeachAugment (TeachA)~\cite{teachaugment}, MADAug~\cite{madaug}, and SoftAug~\cite{softaug}.

\subsubsection{Implementation Details}
We closely follow previous works~\cite{cutout,advmask,trivialaugment} with our setup. Specifically, images are preprocessed by dividing each pixel value by 255 and normalized by the dataset statistics.
We train 1800 epochs with cosine learning rate decay for Shake-Shake~\cite{shake-shake} using SGD with Nesterov Momentum and a learning rate of 0.1, a batch size of 256, 1$e^{-3}$ weight decay and cosine learning rate decay.
We train all other models for 300 epochs with a batch size of 256 and a 0.1 learning rate with cosine annealing learning rate decay strategy, SGD optimizer with the momentum of 0.9, and weight decay of 5$e^{-4}$. 
We follow the common practice in the field of DA method studies~\cite{cutout, advmask} to build the baseline model, i.e., data augmentation of random crop and random horizontal flip is utilized as the baseline. 
For fairness, all methods are implemented with the same configurations.
The augmentation space utilized follows previous work~\cite{autoaugment,trivialaugment,selectaugment}, while the magnitude of each strategy is dynamically determined.
The experiments are repeated across three independent runs.
\subsection{Results on CIFAR-10 and CIFAR-100}
\begin{table}[t]
\caption{Image classification accuracy (\%) on CIFAR-10\slash 100 (average $\pm$ std). * means results reported in the original paper. EA+EL: EntAugment+EntLoss.}
    \label{cifar}
	\centering
	\renewcommand\arraystretch{.9}
	\resizebox{0.97\textwidth}{!}{
			\begin{tabular}{l|lllll|lllll}
				\toprule[1.5pt]
   \multirow{2}{*}{Method} & R-18~\cite{resnet} & R-44~\cite{resnet} & R-50~\cite{resnet} 
   & WRN~\cite{wrn} & SS-32~\cite{shake-shake} & R-18~\cite{resnet} & R-44~\cite{resnet} & R-50~\cite{resnet} 
   & WRN~\cite{wrn} & SS-32~\cite{shake-shake} \\ \cline{2-11}
   & \multicolumn{5}{c|}{CIFAR-10~\cite{cifar}}& \multicolumn{5}{c}{CIFAR-100~\cite{cifar}}\\ \hline
    baseline &95.28\scriptsize{$\pm$.14}*&94.10\scriptsize{$\pm$.40} & 95.66\scriptsize{$\pm$.08}&95.52\scriptsize{$\pm$.11} & 94.90\scriptsize{$\pm$.07}*&77.54\scriptsize{$\pm$.19}*&74.80\scriptsize{$\pm$.38}*&77.41\scriptsize{$\pm$.27}*&78.96\scriptsize{$\pm$.25}*&76.65\scriptsize{$\pm$.14}*\\    HaS~\cite{has}&96.10\scriptsize{$\pm$.14}*&94.97\scriptsize{$\pm$.27}&95.60\scriptsize{$\pm$.15}&96.94\scriptsize{$\pm$.08}&96.89\scriptsize{$\pm$.10}*&78.19\scriptsize{$\pm$.23}&75.82\scriptsize{$\pm$.32}&78.76\scriptsize{$\pm$.24}&80.22\scriptsize{$\pm$.16}&76.89\scriptsize{$\pm$.33}  \\
    Cutout~\cite{cutout}&96.01\scriptsize{$\pm$.18}*&94.78\scriptsize{$\pm$.35}& 95.81\scriptsize{$\pm$.17}&96.92\scriptsize{$\pm$.09}&96.96\scriptsize{$\pm$.09}*&78.04\scriptsize{$\pm$.10}*&74.84\scriptsize{$\pm$.56}&78.62\scriptsize{$\pm$.25}&79.84\scriptsize{$\pm$.14}&77.37\scriptsize{$\pm$.28}  \\
    CutMix~\cite{cutmix}&96.64\scriptsize{$\pm$.22}*&95.28\scriptsize{$\pm$.16}&96.81\scriptsize{$\pm$.10}*&96.93\scriptsize{$\pm$.10}*&96.47\scriptsize{$\pm$.07}&79.45\scriptsize{$\pm$.17}&76.09\scriptsize{$\pm$.15}&81.24\scriptsize{$\pm$.14}&82.67\scriptsize{$\pm$.22}&79.57\scriptsize{$\pm$.10}\\ 
    GridMask~\cite{gridmask}&96.38\scriptsize{$\pm$.17}&95.02\scriptsize{$\pm$.26}&96.15\scriptsize{$\pm$.19}& 96.92\scriptsize{$\pm$.09}&96.91\scriptsize{$\pm$.12}&75.23\scriptsize{$\pm$.21}&76.07\scriptsize{$\pm$.18}&78.38\scriptsize{$\pm$.22}&80.40\scriptsize{$\pm$.20}& 77.28\scriptsize{$\pm$.38} \\
    AdvMask~\cite{advmask}&96.44\scriptsize{$\pm$.15}*&95.49\scriptsize{$\pm$.17}*&96.69\scriptsize{$\pm$.10}*&97.02\scriptsize{$\pm$.05}*&97.03\scriptsize{$\pm$.12}*&78.43\scriptsize{$\pm$.18}*&76.44\scriptsize{$\pm$.18}*&78.99\scriptsize{$\pm$.31}*&80.70\scriptsize{$\pm$.25}*& 79.96\scriptsize{$\pm$.27}* \\
    RE~\cite{randomerasing} &95.69\scriptsize{$\pm$.10}*&94.87\scriptsize{$\pm$.16}*&95.82\scriptsize{$\pm$.17}& 96.92\scriptsize{$\pm$.09}&96.46\scriptsize{$\pm$.13}*&75.97\scriptsize{$\pm$.11}*&75.71\scriptsize{$\pm$.25}*&77.79\scriptsize{$\pm$.32}&80.57\scriptsize{$\pm$.15}&77.30\scriptsize{$\pm$.18} \\
    AA~\cite{autoaugment}&96.51\scriptsize{$\pm$.10}*&95.01\scriptsize{$\pm$.11}&96.59\scriptsize{$\pm$.04}*&96.99\scriptsize{$\pm$.06}&97.30\scriptsize{$\pm$.11}&79.38\scriptsize{$\pm$.20}&76.36\scriptsize{$\pm$.22}&81.34\scriptsize{$\pm$.29}&82.21\scriptsize{$\pm$.17}&82.19\scriptsize{$\pm$.19} \\
    FAA~\cite{fast-autoaugment}&95.99\scriptsize{$\pm$.13}&93.80\scriptsize{$\pm$.12}&96.69\scriptsize{$\pm$.16}&97.30\scriptsize{$\pm$.24}&96.42\scriptsize{$\pm$.12}&79.11\scriptsize{$\pm$.09}&76.04\scriptsize{$\pm$.28}&79.08\scriptsize{$\pm$.12}&79.95\scriptsize{$\pm$.12}&81.39\scriptsize{$\pm$.16} \\
    RA~\cite{randaugment}&96.47\scriptsize{$\pm$.32}&94.38\scriptsize{$\pm$.22}&96.25\scriptsize{$\pm$.06}&96.94\scriptsize{$\pm$.13}*&97.05\scriptsize{$\pm$.15}&78.30\scriptsize{$\pm$.15}&76.30\scriptsize{$\pm$.16}&80.95\scriptsize{$\pm$.22}&82.90\scriptsize{$\pm$.29}*&80.00\scriptsize{$\pm$.29} \\
    
    DADA~\cite{dada}&95.58\scriptsize{$\pm$.06}&93.96\scriptsize{$\pm$.38}&95.61\scriptsize{$\pm$.14}&97.30\scriptsize{$\pm$.13}*&97.30\scriptsize{$\pm$.14}*&78.28\scriptsize{$\pm$.22}&74.37\scriptsize{$\pm$.47}&80.25\scriptsize{$\pm$.28}&82.50\scriptsize{$\pm$.26}*& 80.98\scriptsize{$\pm$.15}\\ 
    TA~\cite{trivialaugment}&96.28\scriptsize{$\pm$.10}&95.00\scriptsize{$\pm$.10}&97.13\scriptsize{$\pm$.08}&97.18\scriptsize{$\pm$.11}&97.30\scriptsize{$\pm$.10}&78.67\scriptsize{$\pm$.19}&76.57\scriptsize{$\pm$.14}&81.34\scriptsize{$\pm$.18}&82.75\scriptsize{$\pm$.26}&82.14\scriptsize{$\pm$.16} \\

    TeachA~\cite{teachaugment} &96.47\scriptsize{$\pm$.13}&95.05\scriptsize{$\pm$.21}&96.40\scriptsize{$\pm$.14}&97.50\scriptsize{$\pm$.16}&97.29\scriptsize{$\pm$.11}&79.27\scriptsize{$\pm$.24}&76.18\scriptsize{$\pm$.31}&80.54\scriptsize{$\pm$.25}&82.81\scriptsize{$\pm$.26}&81.30\scriptsize{$\pm$.18} \\
    MADAug~\cite{madaug}&96.49\scriptsize{$\pm$.12}&95.25\scriptsize{$\pm$.18}&97.12\scriptsize{$\pm$.17}&97.48\scriptsize{$\pm$.15}&97.37\scriptsize{$\pm$.11}&79.39\scriptsize{$\pm$.18}&76.49\scriptsize{$\pm$.21}&81.40\scriptsize{$\pm$.18}&83.01\scriptsize{$\pm$.23}&81.67\scriptsize{$\pm$.19} \\
    SoftAug~\cite{softaug}&96.43\scriptsize{$\pm$.15}&94.51\scriptsize{$\pm$.20}&96.99\scriptsize{$\pm$.14}&97.15\scriptsize{$\pm$.16}&97.22\scriptsize{$\pm$.19}&79.01\scriptsize{$\pm$.21}&76.41\scriptsize{$\pm$.33}&80.94\scriptsize{$\pm$.33}&82.61\scriptsize{$\pm$.24}&80.33\scriptsize{$\pm$.20} \\ \hline
    \textBF{EntAugment}&96.71\scriptsize{$\pm$.05}&\textBF{95.76}\scriptsize{$\pm$.09}&97.09\scriptsize{$\pm$.09}&97.47\scriptsize{$\pm$.10}&97.46\scriptsize{$\pm$.11}&79.45\scriptsize{$\pm$.17}&76.40\scriptsize{$\pm$.18}&81.56\scriptsize{$\pm$.21}&83.09\scriptsize{$\pm$.22} & 81.60\scriptsize{$\pm$.13} \\
    \textBF{EA+EL}&\textBF{96.84}\scriptsize{$\pm$.09}&95.58\scriptsize{$\pm$.03}&\textBF{97.15}\scriptsize{$\pm$.09}&\textBF{97.70}\scriptsize{$\pm$.12}&\textBF{97.55}\scriptsize{$\pm$.10}&\textBF{79.82}\scriptsize{$\pm$.12}&\textBF{76.84}\scriptsize{$\pm$.03}&\textBF{82.49}\scriptsize{$\pm$.15}&\textBF{83.16}\scriptsize{$\pm$.23}&\textBF{82.29}\scriptsize{$\pm$.18}  \\ 
%     \hline
%     \multicolumn{6}{c}{CIFAR-100~\cite{cifar}} \\ \hline
%      baseline &77.54*&74.80*&77.41*&78.96*&76.65* \\
%     HaS~\cite{has}&78.19&75.82&78.76&80.22&76.89 \\
%     Cutout~\cite{cutout}&78.04*&74,84&78.62&79.84&77.37 \\
%     GridMask~\cite{gridmask}&75.23&76.07&78.38&80.40& 77.28\\
%     AdvMask~\cite{advmask}&78.43*&76.44*&78.99*&80.70*& 79.96*\\
%     RandomErasing~\cite{randomerasing} &75.97*&75.71*&77.79&80.57&77.30 \\
%     AutoAugment~\cite{autoaugment}&79.38&76.36&81.34&82.21&82.19 \\
%     FAA~\cite{fast-autoaugment}&79.11&76.04&79.08&79.95&81.39 \\
%     RandAugment~\cite{randaugment}&78.30&76.30&80.95&82.90*&80.00 \\
%     TrivialAugment~\cite{trivialaugment}&78.67&76.80&81.34&82.75&82.14 \\
%     EntAugment&79.27&76.40&81.56&83.09 & 81.60\\
% EntAugment+EntLoss&\textBF{79.82}&\textBF{76.84}&\textBF{82.49}&\textBF{83.16}&\textBF{82.29} \\
        \bottomrule[1.5pt]
		\end{tabular}}
	
\end{table}
To evaluate the effectiveness of EntAugment and EntLoss, in Table~\ref{cifar}, we conduct experiments on CIFAR-10\slash 100 utlizing various deep networks, including ResNet18\slash 44\slash 50 (R-18\slash 44\slash 50)~\cite{resnet}, Wide-ResNet-28-10 (WRN)~\cite{wrn}, and Shake-Shake-26-32 (SS-32)~\cite{shake-shake}.
It can be observed that when employed independently, EntAugment consistently surpasses prior SOTA methods in most cases.
Moreover, the integration of EntLoss into the EntAugment framework consistently demonstrates notable performance improvements across deep models on both CIFAR-10 and CIFAR-100 datasets.
For instance, on CIFAR-10, despite the high test accuracy already achieved by DA methods, both EntAugment and EntAugment+EntLoss further enhance the model performance.
% even when DA methods have already achieved a high level of test accuracy, both EntAugment+EntLoss and EntAugment alone further enhance the performance of deep networks.
In particular, EntAugment+EntLoss typically yields accuracy improvements of approximately 0.5\% on CIFAR-10.
On CIFAR-100, the improvements are even more substantial, with the proposed method surpassing previous SOTA methods by nearly 1\% when employing ResNet-18\slash 50 architectures.
Hence, the proposed adaptive DA framework is more effective in boosting model performance.

Meanwhile, it is worth noting that such substantial performance improvements are achieved without incurring noticeable computational costs, underscoring its superior efficiency.
Consequently, EntAugment and EntLoss can serve as highly efficient plug-and-play techniques for model training.

\begin{figure}[t]
	\centering
         \includegraphics[width=11cm]{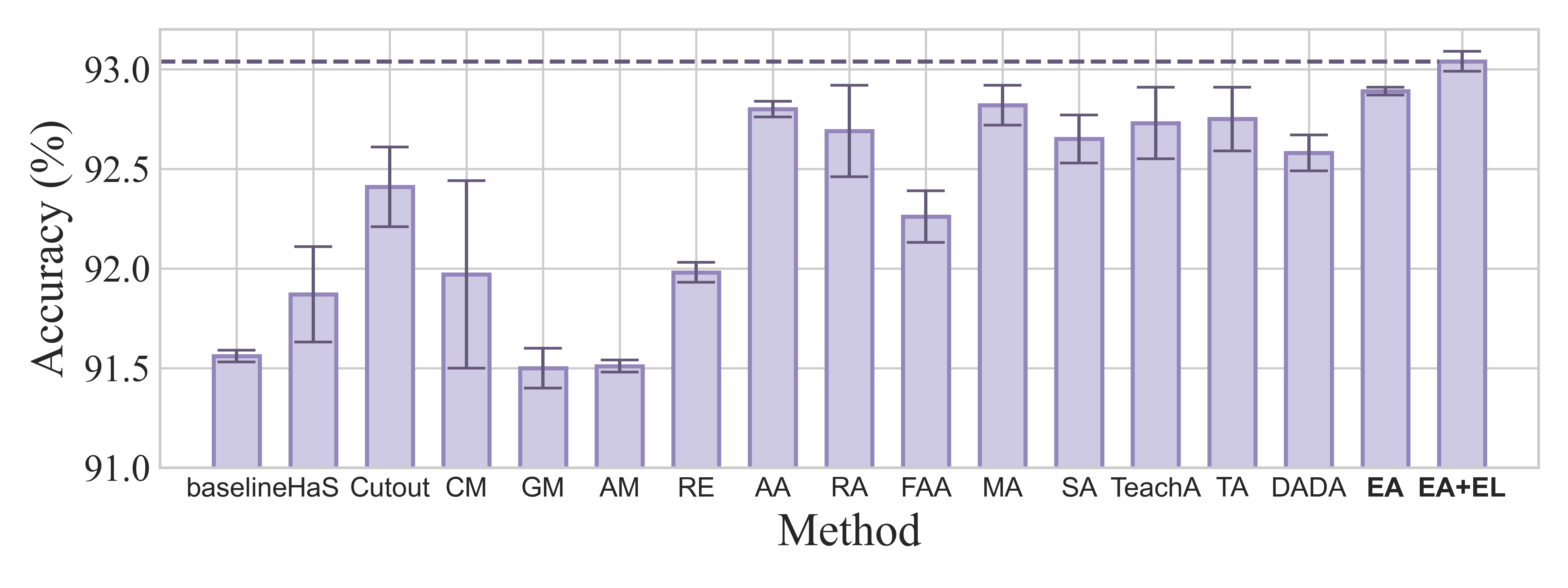}
	\caption{\textBF{Transferalbility analysis.} Transferred test accuracy (\%) on CIFAR-10.}
    \label{fig:transfer-acc}
\end{figure}
\subsection{Transfer Learning}
In the realm of data augmentation, transfer learning is often utilized to assess the transferability of DA methods~\cite{transfer-learning-1,transfer-learning-2,hou2023learn}.
Thus, we pre-train models on the CIFAR-100 dataset using various augmentations and fine-tune these models on CIFAR-10 using ResNet-50.
Theoretically, models trained using more effective DA methods should yield stronger transferability.
As shown in Figure~\ref{fig:transfer-acc}, while the discrepancies in transferred accuracy may appear subtle, it can be observed that EntAugment outperforms other SOTA methods in terms of transferred test accuracy.
Moreover, EntAugment+EntLoss demonstrates even more significant performance improvements.
\subsection{Results on Large-scale ImageNet}
\begin{table}[t]
     \centering
         \caption{Top-1 accuracy (\%) on ImageNet dataset (average $\pm$ std).}\label{tab:imagenet}
	\renewcommand\arraystretch{1.2}
	\resizebox{1.\linewidth}{!}{
			\begin{tabular}{ccccccccccccccc}
				\toprule[1.pt]
				 HaS&GM&Cutout&CutMix& Mixup&AA & FAA & RA &MA&SA&DADA&TA & TeachA 
    &\textBF{EA} &\textBF{EA+EL}   \\ \hline
    77.2\scriptsize{$\pm$0.2}& 77.9\scriptsize{$\pm$0.2} & 77.1\scriptsize{$\pm$0.3} &  77.2\scriptsize{$\pm$0.2} & 77.0\scriptsize{$\pm$0.2}& 77.6\scriptsize{$\pm$0.2}&77.6\scriptsize{$\pm$0.2}&77.6\scriptsize{$\pm$0.2}&\textBF{78.5}\scriptsize{$\pm$0.1}&78.0\scriptsize{$\pm$0.1}&77.5\scriptsize{$\pm$0.1}&77.9\scriptsize{$\pm$0.3}&77.8\scriptsize{$\pm$0.2}
    &78.2\scriptsize{$\pm$0.1}
    &78.3\scriptsize{$\pm$0.1} \\
    \bottomrule[1.pt]
    \end{tabular}}
\end{table}
We also evaluate our framework on the large-scale ImageNet~\cite{imagenet} dataset.
Specifically, we train ResNet-50 models on ImageNet using various DA methods, closely following the experimental setup in ~\cite{autoaugment,trivialaugment}.
In Table~\ref{tab:imagenet}, it can be observed that while prior methods present similar and modest improvements compared to the baseline (e.g., less than 0.9\%), our methods demonstrate a substantial superiority by achieving improvements exceeding 1.2\%.
While EA is slightly worse than MADAug, ours achieves a 2x speed increase over MADAug.
Meanwhile, the training of our proposed methods achieves a 4x faster than TeachAugment.
This highlights the superior effectiveness of our approaches on large-scale datasets. 

Notably, most well-performed methods (e.g., AdvMask~\cite{advmask}, AutoAugment~\cite{autoaugment}, RandAugment~\cite{randaugment}, TeachAugment~\cite{teachaugment}, MADAugment~\cite{madaug} etc.) entail significant additional training costs.
Conversely, the proposed methods obtain nearly equivalent computational overhead compared to the baseline.
Consequently, our proposed methods achieve better results while incurring nearly negligible additional computational overhead.
More results on Tiny-Imagenet and large-scale long-tailed ImageNet-LT~\cite{lt-datasets} and Places-LT~\cite{lt-datasets} are provided in Appendix D and E.
\subsection{t-SNE Visualizations}
\begin{figure}[t]
	\centering
	\begin{subfigure}[]{0.24\linewidth}
		\centering
		\includegraphics[width=3cm]{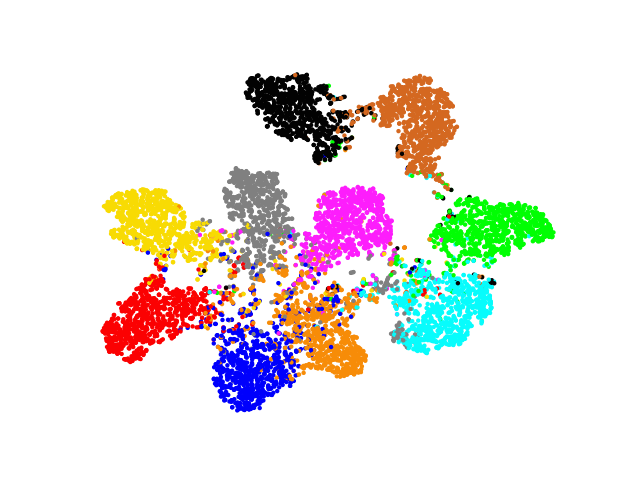}
		\caption{bseline\\ DI=5.03$\times 10^{-5}$}
		\label{fig2-1}
	\end{subfigure}
        \begin{subfigure}[]{0.24\linewidth}
		\centering
		\includegraphics[width=3cm]{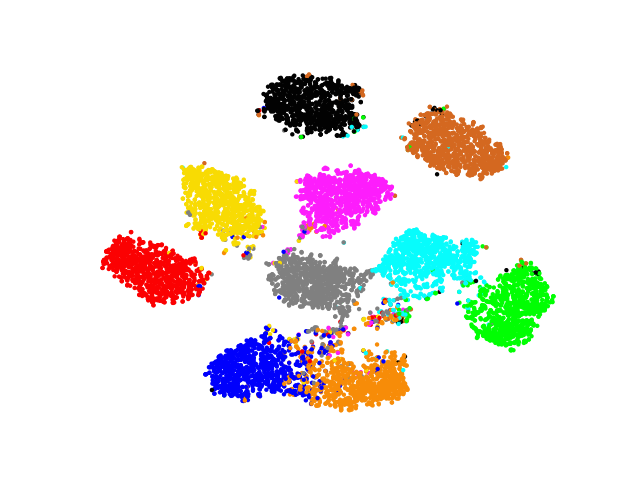}
		\caption{EntAugment\\ DI=7.93$\times 10^{-5}$}
		\label{fig2-2}
	\end{subfigure}
	\begin{subfigure}[]{0.24\linewidth}
		\centering
		\includegraphics[width=3cm]{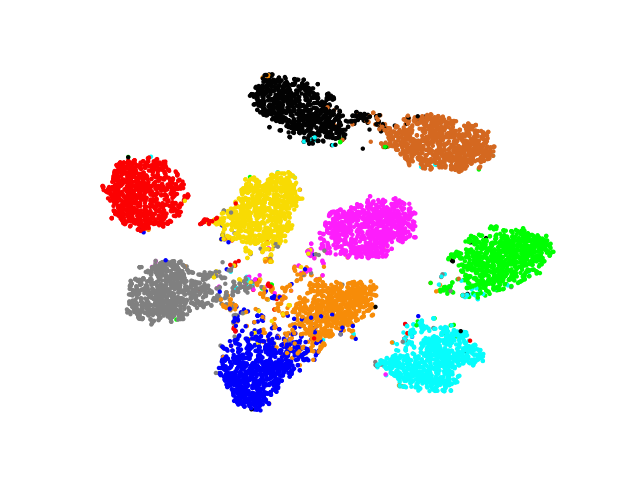}
		\caption{EntLoss\\ DI=8.62$\times 10^{-5}$}
		\label{fig2-3}
	\end{subfigure}
	\begin{subfigure}[]{0.24\linewidth}
		\centering
		\includegraphics[width=3cm]{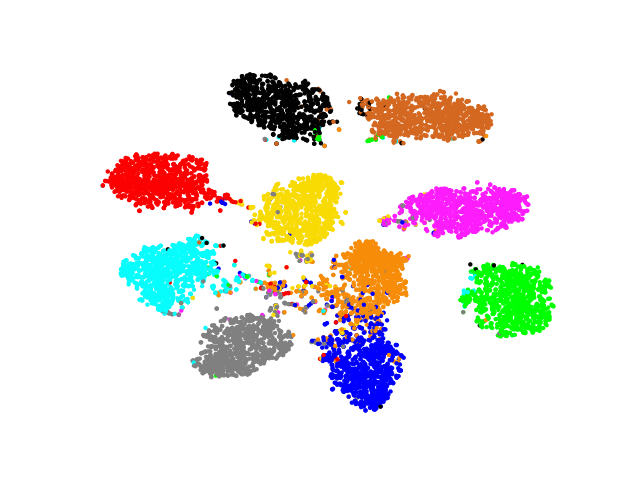}
		\caption{EA+EL \\ DI=1.07$\times 10^{-4}$} 
		\label{fig2-4}
	\end{subfigure}  
	\caption{\textBF{t-SNE Visualization of CIFAR-10 dataset.} DI: Dunn index. EA+EL: EntAugment+EntLoss.}
	\label{fig-tsne}
\end{figure}
To conduct a comparative analysis of model performance with and without utilizing the proposed methods, we visualize the deep features of the CIFAR-10 dataset using t-SNE algorithm~\cite{tsne}.
Specifically, we employ EntAugment, EntLoss, and EntAugment+EntLoss to train ResNet-18 models.
Subsequently, we extract deep features using these trained models and leverage the t-SNE algorithm to analyze the model performance.

In Figure~\ref{fig-tsne}, it can be observed intuitively that applying EntAugment or EntLoss independently results in a pronounced delineation in cluster distribution, i.e., \textBF{enhanced inter-cluster separation and intra-cluster compactness}.
This phenomenon is further enhanced when EntAugment and EntLoss are utilized jointly, as shown in Figure~\ref{fig-tsne}\subref{fig2-4}, where the discriminative characteristics of learned features are effectively enhanced.
Therefore, the proposed methods can bolster the generalization capabilities of the models, thereby facilitating the extraction of more discriminative features.

Moreover, we utilize the Dunn index (DI)~\cite{dunn} to quantitatively analyze the clustering results, which is $DI = \min _{1 \leq i \neq j \leq m} \delta\left(C_i, C_j\right) / \max _{1 \leq j \leq m} \Delta_j$, where separation $\delta\left(C_i, C_j\right)$ is the inter-cluster distance metric between clusters $C_i$ and $C_j$, and compactness $\Delta_j$ calculates the mean distance between all pairs in each cluster.
Hence, a higher DI means better clustering. 
As shown in Figure~\ref{fig-tsne}, using EntAugment and EntLoss individually can bring higher DI values than the baseline.
More precisely, the DI values achieved by EntAugment and EntLoss are found to be 57.7\% and 71.4\% higher than those of the baseline, respectively.
Furthermore, for EntAugment $+$ EntLoss, the DI value experiences a substantial increase, effectively doubling the DI value of the baseline.
Consequently, EntAugment and EntLoss can be utilized separately or jointly to enhance the model performance. 
This investigation not only underscores the efficacy of these methods but also contributes to the interpretability of the proposed techniques.

% The quantitative analysis aligns coherently with the qualitative assessment, thus validating the effectiveness of EntAugment and EntLoss in improving model performance.

\begin{figure}[t]
	\centering
         \includegraphics[width=9cm]{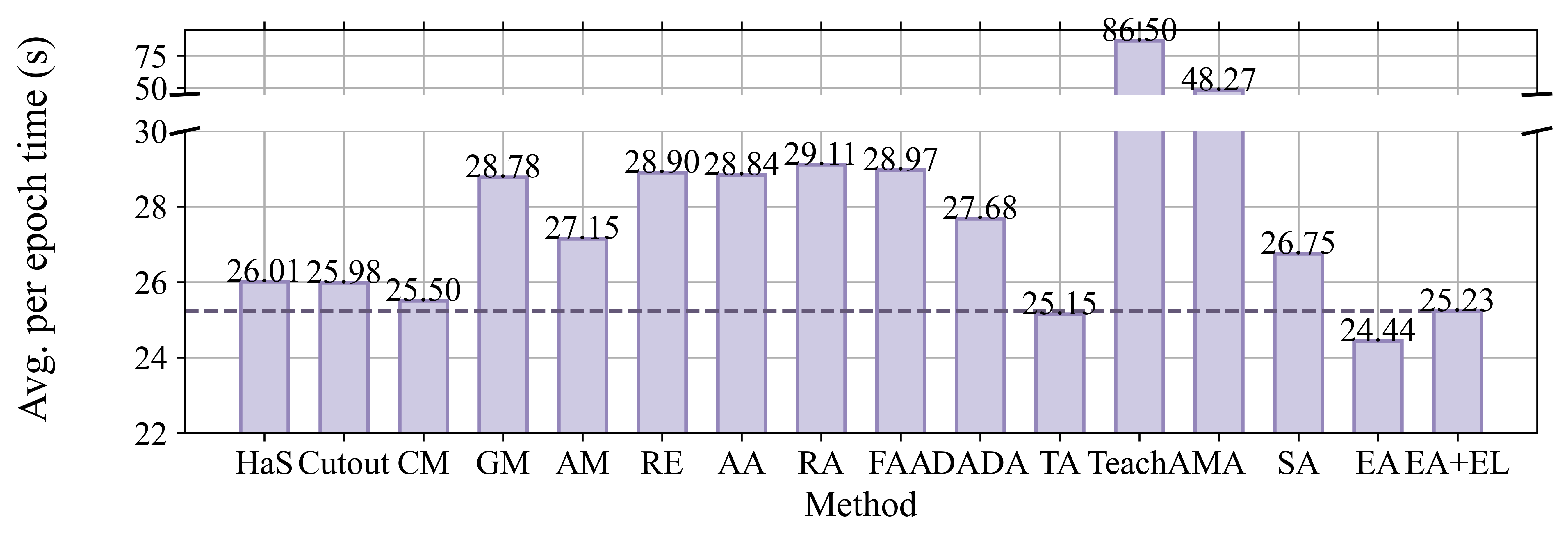}
	\caption{\textBF{Time consumption.} Comparison of training cost of various DA methods.}
    \label{fig:training-cost}
\end{figure}

\subsection{Comparison of Training Efficiency}
To demonstrate the efficiency of our proposed methods, we present a comparison of the training cost associated with employing various DA methods.
All experiments are conducted on 2 NVIDIA RTX2080TI GPUs with batch size 128 and 8 parallel workers.
The experiments are repeated across five independent runs.
The average time cost per epoch is presented in Figure~\ref{fig:training-cost}.
Consistent with the theoretical analysis in Section~\ref{sec:time_cost}, the time cost of our proposed methods remains at \textBF{the lowest level}, highlighting the efficiency.
Although the proposed methods obtain similar time costs with CutMix and TrivialAugment, ours consistently achieves better performance, highlighting the practical effectiveness.
 
\subsection{Convergence Analysis}
To more clearly present the dynamic evolution of test errors throughout the training process, we train ResNet-110 models~\cite{resnet} on CIFAR-10 using a multi-step learning rate decay schedule. The learning rate is initialized as 0.1 and multiplied by 0.2 at epochs 60, 120, 160, 220, and 280.
It is worth noting that all other experimental settings remain unchanged.
The convergence trajectory is shown in Figure~\ref{fig-convergence}.
It can be observed that both EntAugment and EntLoss achieve a significant improvement after the second learning rate drop.
Meanwhile, when EntAugment is combined with EntLoss, it presents lower error rates, suggesting more efficient convergence.

% \begin{figure}[]
% 	\centering
%          \includegraphics[width=6cm]{./pictures/convergence/test.pdf}
% 	\caption{Curves of test errors on CIFAR-10 with ResNet-110.}
% 	\label{fig-convergence}
% \end{figure}

\subsection{Ablation Study}
% \textbf{The effect of EntAugment}
\subsubsection{The effect of EntAugment}
To validate the efficacy of EntAugment, we present the results of merely using EntAugment across benchmark datasets in Table~\ref{cifar} and \ref{tab:imagenet}, as well as results in Appendix D, F, and H.
The results demonstrate that the EntAugment framework outperforms prior SOTA DA methods in most cases, highlighting the effectiveness of the adaptive DA scheme.
\subsubsection{The effect of EntLoss}
\begin{figure}[t]
    \centering
    \begin{minipage}{0.4\textwidth}
         \includegraphics[width=1.\textwidth]{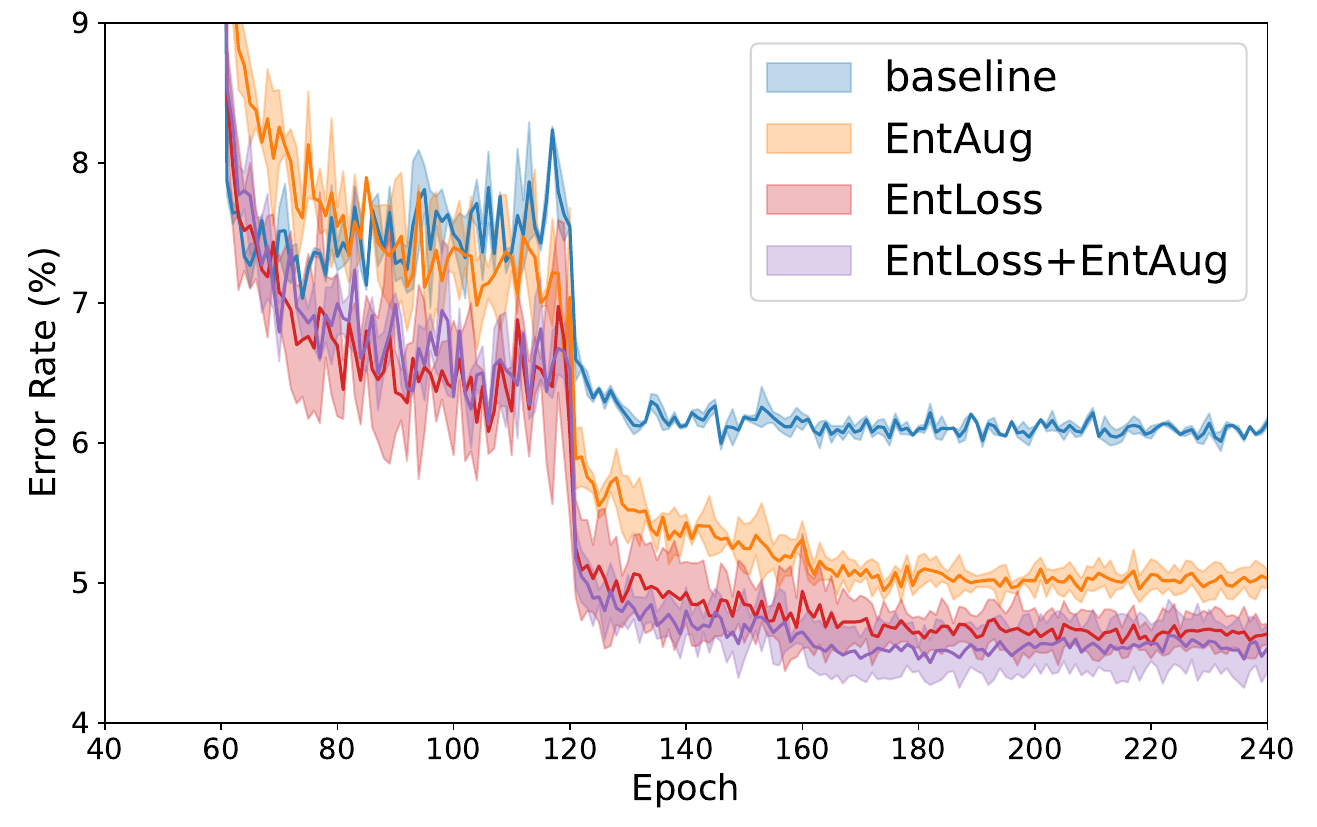}
	\caption{\textBF{Convergence analysis.} Test error on CIFAR with ResNet-110.}
	\label{fig-convergence}
    \end{minipage}
 \begin{minipage}{0.59\textwidth}
     \captionof{table}{\textBF{Effect of EntLoss.} Comparative analysis between the test accuracy with and without employing EntLoss. CELoss: cross-entropy loss.}
    \label{entloss}
	\centering
	\renewcommand\arraystretch{.9}
	\resizebox{1.\textwidth}{!}{
		%		\resizebox{.7\textwidth}{!}{
        \begin{tabular}{l|cc|cc}
            \toprule[1.pt]
            model&CELoss&CELoss+\textBF{EntLoss} &CELoss&CELoss+\textBF{EntLoss} \\ \hline
           &\multicolumn{2}{c|}{CIFAR-10~\cite{cifar}}&\multicolumn{2}{c}{CIFAR-100~\cite{cifar}} \\ \hline
           ResNet18~\cite{resnet} &95.28\%&\textBF{95.41\%} \scriptsize{$\uparrow$0.13}&77.54\%&\textBF{79.17\%} \scriptsize{$\uparrow$1.63} \\
            ResNet44~\cite{resnet} &94.10\%&\textBF{94.73\%} \scriptsize{$\uparrow$0.63}&71.75\%&\textBF{73.20\%} \scriptsize{$\uparrow$1.45} \\
            ResNet50~\cite{resnet}&95.66\%& \textBF{95.88\%} \scriptsize{$\uparrow$0.22}&77.41\%& \textBF{80.29\%} \scriptsize{$\uparrow$2.88}\\
            WRN-28-10~\cite{wrn}&95.52\%&\textBF{96.28\%} \scriptsize{$\uparrow$0.76}&78.96\%&\textBF{80.78\%} \scriptsize{$\uparrow$1.82}\\
            Shake-26-32~\cite{shake-shake}&94.90\%&\textBF{96.55\%} \scriptsize{$\uparrow$0.65}&76.65\%&\textBF{79.15\%} \scriptsize{$\uparrow$2.50} \\  
    \bottomrule[1.pt]
    \end{tabular}}
 \end{minipage}
\end{figure}

In this section, we will demonstrate that the utilization of EntLoss yields substantial improvements compared to conventional classification model training.
Comparative analyses between models trained with and without utilizing EntLoss are presented in Table~\ref{entloss}. 
It can be observed that EntLoss brings a notable improvement in model performance when combined with vanilla CE loss (consistent with the theoretical analysis in Section~\ref{sec:entloss}).
% effectively improves the performance of vanilla CE loss across various popular deep models.
Especially on CIFAR-100, EntLoss can improve test accuracy by nearly 2\%.
It is also noteworthy that such significant improvements are achieved without introducing any noticeable computational overhead.
%  \begin{table}[t]
%  \caption{The comparative analysis between the test accuracy with and without employing EntLoss. CELoss: cross-entropy loss.}
%     \label{entloss}
% 	\centering
% 	\renewcommand\arraystretch{.8}
% 	\resizebox{0.7\textwidth}{!}{
% 		%		\resizebox{.7\textwidth}{!}{
%         \begin{tabular}{l|cc|cc}
%             \toprule[1.pt]
%             model&CELoss&CELoss+EntLoss &CELoss&CELoss+EntLoss \\ \hline
%            &\multicolumn{2}{c|}{CIFAR-10~\cite{cifar}}&\multicolumn{2}{c}{CIFAR-10~\cite{cifar}} \\ \hline
%            ResNet18~\cite{resnet} &95.28\%&\textBF{95.41\%} \scriptsize{$\uparrow$0.13}&77.54\%&\textBF{79.17\%} \scriptsize{$\uparrow$1.63} \\
%             ResNet44~\cite{resnet} &94.10\%&\textBF{94.73\%} \scriptsize{$\uparrow$0.63}&71.75\%&\textBF{73.20\%} \scriptsize{$\uparrow$1.45} \\
%             ResNet50~\cite{resnet}&95.66\%& \textBF{95.88\%} \scriptsize{$\uparrow$0.22}&77.41\%& \textBF{80.29\%} \scriptsize{$\uparrow$2.88}\\
%             WRN-28-10~\cite{wrn}&95.52\%&\textBF{96.28\%} \scriptsize{$\uparrow$0.76}&78.96\%&\textBF{80.78\%} \scriptsize{$\uparrow$1.82}\\
%             Shake-26-32~\cite{shake-shake}&94.90\%&\textBF{96.55\%} \scriptsize{$\uparrow$0.65}&76.65\%&\textBF{79.15\%} \scriptsize{$\uparrow$2.50} \\  
%     \bottomrule[1.pt]
%     \end{tabular}}
% \end{table}

\section{Conclusion}
% In this paper, we point out that augmenting data using DA operations with random magnitudes within the current DA framework may result in excessive or insufficient data manipulation, leading to noisy samples, distribution shifts, and elevated risk of overfitting.
In this paper, we propose a novel, tuning-free, and adaptive DA framework, EntAugment.
EntAugment operates without the need for manual tuning and dynamically adjusts the magnitudes of DA applied to each training data during online training based on sample difficulty and model training progression.
Moreover, we also introduce an entropy regularization loss, EntLoss, to enhance the effectiveness of EntAugment for better generalization.
Through theoretical analysis, we show that EntLoss brings more significant benefits of achieving closer alignment between the model's distribution and the dataset's inherent distribution.
Notably, without introducing auxiliary models or extra training overhead, both EntAugment and EntLoss introduce minimal computational cost to the task model training process, ensuring ease of integration and practical feasibility. 
% Thus, the proposed methods are easy to use and efficacious for enhancing the generalization performance of deep networks.
Experimental results on several benchmark datasets show that the proposed methods achieve state-of-the-art performance while showcasing superior efficiency.
Meanwhile, models trained with EntAugment can exhibit enhanced transferability and generalization capabilities compared to other augmentations.
% superior performance compared to state-of-the-art while exhibiting exceptional efficiency.
In the future, we will explore the application of the proposed methods on other popular tasks, e.g., self-supervised learning and object detection.

\section*{Acknowledgements}
This work was supported in part by the STI 2030-Major Projects of China under Grant 2021ZD0201300 and by the National Natural Science Foundation of China under Grant 62276127.

\bibliographystyle{splncs04}
\bibliography{main}

\clearpage
\setcounter{page}{1}
\appendix
\section{The Augmentation Space of EntAugment}
 \begin{table}[]
 \centering
        \caption{The Augmentation Space.}
 \renewcommand\arraystretch{1.}
 \resizebox{.32\columnwidth}{!}
 {
  \begin{tabular}{c|c|c}
   \toprule[1.2pt]
           Transformation & $S_{Max}$ & Symmetric  \\
           \hline
           identity & - & - \\
           auto contrast & - & - \\
           equalize & - & - \\
           color & 1.9 & - \\
           contrast & 1.9 & - \\
           brightness & 1.9 & - \\
           sharpness & 1.9 & - \\
           rotation & $30^\circ$ & $\pm$ \\
           $\text{translate}_x$ &10 & $\pm$ \\
           $\text{translate}_y$ &10 & $\pm$ \\
          $\text{shear}_x$ & 0.3 & $\pm$ \\
           $\text{shear}_y$ & 0.3 & $\pm$  \\
            solarize & 256 &- \\
            posterize & 4 & - \\
           
   \bottomrule[1.2pt]
  \end{tabular}
  }
 \label{tab:augmentation-space}
\end{table}
In Table~\ref{tab:augmentation-space}, we present all the transformations used in EntAugment.
Each transformation has a maximum allowable magnitude $S_{MAX}$.
The applied strength is $S_{MAX} \times m$, where $m \in [0,1]$ is the magnitude value.
For symmetric transformations (e.g., rotation, etc.), the symmetric direction ($\pm$) is selected randomly.

 \section{Proof of Lemma~\ref{lemma1}}~\label{sec:lemma1}
 \begin{proof}
     The goal of maximum likelihood estimation is to find the values of the model parameters $\boldsymbol{\theta}$ that maximize the likelihood function over the parameter space, that is,
    \begin{align}
       \boldsymbol{\hat{\theta}}  &= {\arg \max_{\boldsymbol{\theta} } p_{\text {model }}(\mathcal{D} ; \boldsymbol{\theta})} \\
        &= \arg \max_{\boldsymbol{\theta}} \prod_{i=1}^n p_{\text {model }}\left(\boldsymbol{z}_i ; \boldsymbol{\theta}\right)\\
        &= \arg \max_{\boldsymbol{\theta}} \sum_{i=1}^n \log p_{\text {model }}(\boldsymbol{z}_i;\boldsymbol{\theta}) \\
        &= \arg \max_{\boldsymbol{\theta}} \mathbb{E}_{\mathbf{z} \sim \hat{p}_{\text {data }}} \log p_{\text {model }}(\boldsymbol{z} ; \boldsymbol{\theta}).\label{eq:ML}
    \end{align}
 \end{proof}

\section{Proof of Proposition~\ref{prop1}}~\label{sec:prop1}
\begin{proof}
    According to Eq.~\eqref{eq:KL}, the difference in the approximation level of $\hat{\theta}_{cro}$ and $\hat{\theta}_{ent}$ in fitting empirical dataset distribution can be expressed as:
\begin{align}
    &D_{KL}(\hat{p}_{data} || p_{\hat{\theta}_{cro}}) - D_{KL}(\hat{p}_{data} || p_{\hat{\theta}_{ent}}) \\
    \begin{split}
    =&\mathbb{E}_{\mathbf{z} \sim \hat{p}_{\text {data }}}\left[\log \hat{p}_{\text {data }}(\boldsymbol{z})-\log p_{\hat{\theta}_{cro}}(\boldsymbol{z})\right] \\
    & - \mathbb{E}_{\mathbf{z} \sim \hat{p}_{\text {data }}}\left[\log \hat{p}_{\text {data }}(\boldsymbol{z})-\log p_{\hat{\theta}_{ent}}(\boldsymbol{z})\right]
    \end{split}\\
    =& \mathbb{E}_{\mathbf{z} \sim \hat{p}_{\text {data }}}\left[\log p_{\hat{\theta}_{ent}}(\boldsymbol{z}) - \log p_{\hat{\theta}_{cro}}(\boldsymbol{z})   \right] \\
    = &\mathbb{E}_{\mathbf{z} \sim \hat{p}_{\text {data }}}\left[\log p_{\hat{\theta}_{ent}}(\boldsymbol{z})\right] - \mathbb{E}_{\mathbf{z} \sim \hat{p}_{\text {data }}}\left[\log p_{\hat{\theta}_{cro}}(\boldsymbol{z})\right]\\
    \geq & 0. \label{eq:18}
\end{align}
Inequality~\eqref{eq:18} holds according to Inequality~\eqref{eq:ent}.
\end{proof}
\section{Results on Tiny-ImageNet}\label{sec:tiny}
\begin{table}[]
 \caption{\textBF{Image classification accuracy (\%) on Tiny-ImageNet dataset} (average $\pm$ std). * means results reported in the previous paper. EA+EL: EntAugment+EntLoss. }
		\label{tab:tiny}
	\centering
	\renewcommand\arraystretch{.9}
	\resizebox{0.5\textwidth}{!}{
		%		\resizebox{.7\textwidth}{!}{
			\begin{tabular}{l|lllll}
				\toprule[1.5pt]
				 Method & ResNet-18~\cite{resnet} & ResNet-50~\cite{resnet} & WRN-50-2~\cite{wrn}  \\ \hline
     baseline &61.38\scriptsize{$\pm$0.99}&73.61\scriptsize{$\pm$0.43}&81.55\scriptsize{$\pm$1.24} \\
    HaS~\cite{has} &63.51\scriptsize{$\pm$0.58}&75.32\scriptsize{$\pm$0.59}&81.77\scriptsize{$\pm$1.16} \\
    Cutout~\cite{cutout} &68.67\scriptsize{$\pm$1.06}&77.45\scriptsize{$\pm$0.42}&82.27\scriptsize{$\pm$1.55} \\
    CutMix~\cite{cutmix} &64.09\scriptsize{$\pm$0.30}&76.41\scriptsize{$\pm$0.27}&82.32\scriptsize{$\pm$0.46} \\
    GridMask~\cite{gridmask} &62.72\scriptsize{$\pm$0.91}*&77.88\scriptsize{$\pm$2.50}&82.25\scriptsize{$\pm$1.47}\\
    AdvMask~\cite{advmask} &65.29\scriptsize{$\pm$0.20}*&79.84\scriptsize{$\pm$0.28}*&83.39\scriptsize{$\pm$0.55}*\\
    RandomErasing~\cite{randomerasing} &64.00\scriptsize{$\pm$0.37}&75.33\scriptsize{$\pm$1.58}&81.89\scriptsize{$\pm$1.40} \\
    AutoAugment~\cite{autoaugment} &67.28\scriptsize{$\pm$1.40}&75.29\scriptsize{$\pm$2.40}&79.99\scriptsize{$\pm$2.20} \\
    FAA~\cite{fast-autoaugment} &68.15\scriptsize{$\pm$0.70}&75.11\scriptsize{$\pm$2.70}&82.90\scriptsize{$\pm$0.92} \\
    RandAugment~\cite{randaugment} &65.67\scriptsize{$\pm$1.10}&75.87\scriptsize{$\pm$1.76}&82.25\scriptsize{$\pm$1.02}\\
    DADA~\cite{dada}&70.03\scriptsize{$\pm$0.10}&78.61\scriptsize{$\pm$0.34}&83.03\scriptsize{$\pm$0.18} \\
    TrivialAugment~\cite{trivialaugment} &69.97\scriptsize{$\pm$0.96}&78.98\scriptsize{$\pm$0.39}&82.16\scriptsize{$\pm$0.32}\\ \hline
    \textBF{EntAugment} &70.16\scriptsize{$\pm$0.92}&\textBF{79.06}\scriptsize{$\pm$0.32}&83.92\scriptsize{$\pm$0.24} \\
    \textBF{EA+EL}&\textBF{70.55}\scriptsize{$\pm$1.05}&78.75\scriptsize{$\pm$0.20}&\textBF{84.03}\scriptsize{$\pm$0.27} \\
    \bottomrule[1.5pt]
    \end{tabular}}

\end{table}

 \begin{table}[h]
 \caption{Top-1 classification accuracy (\%) on ImageNet-LT and Places-LT. * means results reported in the original paper. \textBF{EA+EL}: EntAugment+ EntLoss. The performance is average$\pm$std.  }
    \label{tab:tl}
    \centering
    \begin{subtable}{1.\textwidth}
    \caption{Top-1 classification accuracy on ImageNet-LT.}
    \renewcommand\arraystretch{1.}
        \resizebox{1.\textwidth}{!}{\begin{tabular}{c|llll|llll}
          \toprule[1.2pt]
          \textBF{Backbone Net}& \multicolumn{4}{c|}{\textBF{closed-set setting}} &\multicolumn{4}{c}{\textBF{open-set setting}} \\ 
          ResNet-10 & \multicolumn{1}{c}{$> 100$} & \multicolumn{1}{c}{$\leq 100$ \& $>20$} &\multicolumn{1}{c}{$<20$} && \multicolumn{1}{c}{$> 100$} & \multicolumn{1}{c}{$\leq 100$ \& $>20$} &\multicolumn{1}{c}{$<20$}& \\
          \textBF{Methods} &\multicolumn{1}{c}{\textBF{Many-shot}}& \multicolumn{1}{c}{\textBF{Medium-shot}} & \multicolumn{1}{c}{\textBF{Few-shot}} & \multicolumn{1}{c|}{\textBF{Overall}} &\multicolumn{1}{c}{\textBF{Many-shot}} & \multicolumn{1}{c}{\textBF{Medium-shot}} & \multicolumn{1}{c}{\textBF{Few-shot}} & \multicolumn{1}{c}{\textBF{F-measure}} \\ \hline
          OLTR~\cite{lt-datasets} & 43.2\scriptsize{$\pm$0.1}* & 35.1\scriptsize{$\pm$0.2}* & 18.5\scriptsize{$\pm$0.1}* & 35.6\scriptsize{$\pm$0.1}* & 41.9\scriptsize{$\pm$0.1}*& 33.9\scriptsize{$\pm$0.1}*& 17.4\scriptsize{$\pm$0.2}* &44.6\scriptsize{$\pm$0.2}* \\
          OLTR+\textBF{EA} & 45.0\scriptsize{$\pm$0.1} & 38.4\scriptsize{$\pm$0.1} & \textBF{22.4}\scriptsize{$\pm$0.2} & 38.6\scriptsize{$\pm$0.1} & 44.5\scriptsize{$\pm$0.1} & 37.4\scriptsize{$\pm$0.1} & \textBF{21.8}\scriptsize{$\pm$0.2} & 45.2\scriptsize{$\pm$0.2} \\
           OLTR+\textBF{EA+EL} &\textBF{46.7}\scriptsize{$\pm$0.2}&\textBF{39.2}\scriptsize{$\pm$0.1}&21.5\scriptsize{$\pm$0.1}&\textBF{39.6}\scriptsize{$\pm$0.2} &\textBF{45.1}\scriptsize{$\pm$0.2}&\textBF{37.7}\scriptsize{$\pm$0.1}&20.7\scriptsize{$\pm$0.2} & \textBF{46.3}\scriptsize{$\pm$0.1} \\
         \bottomrule[1.2pt]
        \end{tabular}}
    \end{subtable}
   \begin{subtable}{1.\textwidth}
    \caption{Top-1 classification accuracy on Places-LT.}
    \renewcommand\arraystretch{1.}
        \resizebox{1.\textwidth}{!}{\begin{tabular}{c|llll|llll}
          \toprule[1.2pt]
      \textBF{Backbone Net}& \multicolumn{4}{c|}{\textBF{closed-set setting}} &\multicolumn{4}{c}{\textBF{open-set setting}} \\ 
          ResNet-152 & \multicolumn{1}{c}{$> 100$} & \multicolumn{1}{c}{$\leq 100$ \& $>20$} &\multicolumn{1}{c}{$<20$} && \multicolumn{1}{c}{$> 100$} & \multicolumn{1}{c}{$\leq 100$ \& $>20$} &\multicolumn{1}{c}{$<20$}& \\
          \textBF{Methods} &\multicolumn{1}{c}{\textBF{Many-shot}}& \multicolumn{1}{c}{\textBF{Medium-shot}} & \multicolumn{1}{c}{\textBF{Few-shot}} & \multicolumn{1}{c|}{\textBF{Overall}} &\multicolumn{1}{c}{\textBF{Many-shot}} & \multicolumn{1}{c}{\textBF{Medium-shot}} & \multicolumn{1}{c}{\textBF{Few-shot}} & \multicolumn{1}{c}{\textBF{F-measure}} \\ \hline
       OLTR~\cite{lt-datasets} & \textBF{44.7}\scriptsize{$\pm$0.1}* & 37.0\scriptsize{$\pm$0.2}* & 25.3\scriptsize{$\pm$0.1}* & 35.9\scriptsize{$\pm$0.1}* &\textBF{44.6}\scriptsize{$\pm$0.1}*& 36.8\scriptsize{$\pm$0.1}*& 25.2\scriptsize{$\pm$0.2}* & 46.4\scriptsize{$\pm$0.1}*  \\
         OLTR+\textBF{EA} &44.1\scriptsize{$\pm$0.1} & 40.9\scriptsize{$\pm$0.1} & \textBF{29.8}\scriptsize{$\pm$0.2} & 39.2\scriptsize{$\pm$0.0}& 43.7\scriptsize{$\pm$0.1} & 40.6\scriptsize{$\pm$0.1} & 28.5\scriptsize{$\pm$0.2} & 50.0\scriptsize{$\pm$0.1}  \\
         OLTR+\textBF{EA+EL}&44.0\scriptsize{$\pm$0.1}&\textBF{41.1}\scriptsize{$\pm$0.1}&29.7\scriptsize{$\pm$0.2}&\textBF{39.5}\scriptsize{$\pm$0.1}&44.1\scriptsize{$\pm$0.1}&\textBF{41.0}\scriptsize{$\pm$0.1}&\textBF{29.5}\scriptsize{$\pm$0.2} & \textBF{50.5}\scriptsize{$\pm$0.1}\\
         \bottomrule[1.2pt]
        \end{tabular}}
    \end{subtable}
\end{table}
We conduct experiments on Tiny-ImageNet to assess the efficacy of the proposed method across various popular deep networks, including ResNet-18\slash 50~\cite{resnet} and Wide-ResNet-50-2 (WRN-50-2)~\cite{wrn}.
We resize the images to $64\times 64$, initialize the models with ImageNet pre-trained weights, and then fine-tune models employing various DA methods.
As shown in Table~\ref{tab:tiny}, EntAugment and EntLoss consistently yield superior accuracy across various models, 
Remarkably, among all data augmentation methods, EntAugment+EntLoss achieves the highest accuracy improvements,  with increments of 9.17\% for ResNet-18, 5.14\% for ResNet-50, and 2.48\% for WRN-50-2 compared to the baseline, respectively.

Notably, most well-performed methods (e.g., AdvMask~\cite{advmask}, AutoAugment~\cite{autoaugment}, RandAugment~\cite{randaugment}, etc.) entail significant additional training costs.
Conversely, the proposed methods obtain nearly equivalent computational overhead compared to the baseline.
Consequently, our proposed methods achieve better results while incurring nearly negligible additional computational overhead.

\section{Results on Large-scale Long-tailed Datasets}\label{sec:lt-datasets}
While most DA methods have not been tested on large-scale long-tail biased datasets, we strengthen the generality and effectiveness of our method by applying it to the ImageNet-LT and Places-LT datasets~\cite{lt-datasets}.
Specifically, we utilize the codebase provided by OLTR~\cite{lt-datasets} and rigorously follow all the training settings, except for using EntAugment (\textBF{EA}) and EntLoss (\textBF{EL}).
As illustrated in Table~\ref{tab:tl}, our methods significantly enhance the performance of OLTR across all the settings, including many-shot, medium-shot, few-shot, and overall (e.g., achieving over 4\% gains in accuracy) on both datasets.
Meanwhile, performance improvements can be observed in both closed-set and open-set settings.
The advantage is even more profound under the F-measure.
Notably, these advantages are achieved without introducing any noticeable training overhead, thereby highlighting its significance.

\section{Additional Results of EntLoss}\label{sec:add-entloss}

 \begin{table}[]
	\centering
 \caption{Image classification accuracy (\%) of various DA methods on CIFAR-10 with and without employing EntLoss. The performance is average$\pm$std. }
	\renewcommand\arraystretch{.9}
	\resizebox{0.8\textwidth}{!}{
        \begin{tabular}{l|cc|cc}
				\toprule[1.5pt]
		Method	&ResNet-18~\cite{resnet}& ResNet-50~\cite{resnet}&ResNet-18~\cite{resnet}& ResNet-50~\cite{resnet} \\ \hline
  &\multicolumn{2}{c|}{CIFAR-10} & \multicolumn{2}{c}{CIFAR-100} \\ \hline
  
   Cutout~\cite{cutout}&96.01\scriptsize{$\pm$0.18}&95.81\scriptsize{$\pm$0.17}&78.04\scriptsize{$\pm$0.10}&78.62\scriptsize{$\pm$0.25} \\
  Cutout+\textBF{EntLoss}&\textBF{96.50}\scriptsize{$\pm$0.20}&\textBF{96.49}\scriptsize{$\pm$0.15}&\textBF{79.35}\scriptsize{$\pm$0.16}&\textBF{81.30}\scriptsize{$\pm$0.19} \\ \hline 
  RE~\cite{randomerasing}&95.69\scriptsize{$\pm$0.10}&96.04\scriptsize{$\pm$0.17}&75.97\scriptsize{$\pm$0.11}&77.79\scriptsize{$\pm$0.32}\\  
  RE+\textBF{EntLoss} &\textBF{96.08}\scriptsize{$\pm$0.10} &\textBF{96.46}\scriptsize{$\pm$0.12} &\textBF{75.98}\scriptsize{$\pm$0.15}  &\textBF{79.56}\scriptsize{$\pm$0.25}  \\  \hline
AutoAugment~\cite{autoaugment}&95.04\scriptsize{$\pm$0.10}&96.11\scriptsize{$\pm$0.04}&79.78\scriptsize{$\pm$0.20}&82.08\scriptsize{$\pm$0.29}\\
  AA+\textBF{EntLoss}&\textBF{96.89}\scriptsize{$\pm$0.12}&\textBF{97.00}\scriptsize{$\pm$0.15}&\textBF{80.12}\scriptsize{$\pm$0.18}&\textBF{82.54}\scriptsize{$\pm$0.22}\\  \hline
  FAA~\cite{fast-autoaugment}&95.99\scriptsize{$\pm$0.13}&\textBF{96.69}\scriptsize{$\pm$0.16}&79.11\scriptsize{$\pm$0.09}&79.08\scriptsize{$\pm$0.12} \\
  FAA+\textBF{EntLoss}&\textBF{96.01}\scriptsize{$\pm$0.15}&96.58\scriptsize{$\pm$0.19}&\textBF{79.59}\scriptsize{$\pm$0.11}&\textBF{79.70}\scriptsize{$\pm$0.18} \\  \hline
  RandAugment~\cite{randaugment}&96.47\scriptsize{$\pm$0.32}&96.25\scriptsize{$\pm$0.06}&78.30\scriptsize{$\pm$0.15}&80.95\scriptsize{$\pm$0.22} \\ 
  RA+\textBF{EntLoss}&\textBF{96.48}\scriptsize{$\pm$0.26}&\textBF{96.47}\scriptsize{$\pm$0.16}&\textBF{78.65}\scriptsize{$\pm$0.18}&\textBF{81.64}\scriptsize{$\pm$0.14}\\  \hline
  TrivialAugment~\cite{trivialaugment}&96.70\scriptsize{$\pm$0.10}&97.13\scriptsize{$\pm$0.09}&78.67\scriptsize{$\pm$0.17}&81.33\scriptsize{$\pm$0.21} \\ 
  TA+\textBF{EntLoss}&\textBF{96.75}\scriptsize{$\pm$0.11}&\textBF{97.19}\scriptsize{$\pm$0.19}&\textBF{79.48}\scriptsize{$\pm$0.24}&\textBF{81.53}\scriptsize{$\pm$0.19}\\ \hline
\bottomrule[1.5pt]
    \end{tabular}}

		\label{supp:tab}
\end{table}
To further evaluate the effectiveness of EntLoss, we conduct experiments utilizing EntLoss in conjunction with various DA methods.
The results on both CIFAR-10 and CIFAR-100 datasets are summarized in Table~\ref{supp:tab}.
It can be seen that the integration of EntLoss leads to improved generalization performance across various DA methods, which is consistent with the theoretical analyses in Section~\ref{sec:entloss}.
While there are a few instances where the accuracy improvement is slightly marginal, the predominant trend indicates a significant enhancement in model performance. 
For instance, when integrated with TrivialAugment into a ResNet-18 architecture on the CIFAR-100 dataset, EntLoss results in a performance gain of 0.81\%.
Similarly, employing EntLoss alongside Random Erasing within a ResNet-50 model yields a performance boost of 1.77\%. 
Notably, such performance improvements are achieved with minimal additional computational overhead.
\section{Choice of the Base Augmentation}
\begin{table}[]
    \centering
    \caption{Analysis of the choice of base transformations.}
    \resizebox{0.75\textwidth}{!}{
    \begin{tabular}{c|ccccccc}
                \toprule[1.2pt]
    $M$ &2&4&6&8&10&12&14\\ \hline
    Accuracy &96.36\%&96.47\%&96.61\%&96.58\%&96.64\%&96.74\%&96.84\% \\ 
    \bottomrule[1.2pt]
    \end{tabular}}  
    \label{tab:choice_base_aug}
\end{table}
In this section, we analyze the performance by varying the number of transformations (i.e., $M$) in our augmentation space.
It can be seen in Table~\ref{tab:choice_base_aug} that while performance decreases as fewer augmentations are used, it drops very slowly.
This highlights the effectiveness of EntAugment.

\section{Results on Fine-grained Datasets}
 \begin{table}[]
	\centering
 \caption{Test accuracy (\%) of various DA methods on fine-grained datasets. The performance is average$\pm$std.}
	\renewcommand\arraystretch{1.}
	\resizebox{0.6\textwidth}{!}{
        \begin{tabular}{l|c c c}\toprule[1.5pt]
    Dataset & baseline & EntAugment & EA+EL \\ \hline
    Oxford Flowers~\cite{oxford-flower} &89.47\scriptsize{$\pm$0.08}&97.13\scriptsize{$\pm$0.09}&\textBF{97.19}\scriptsize{$\pm$0.08} \\ 
    Oxford-IIIT Pets~\cite{oxford-pets} &89.73\scriptsize{$\pm$0.18} &91.61\scriptsize{$\pm$0.03} &\textBF{91.80}\scriptsize{$\pm$0.27} \\
    FGVC-Aircraft~\cite{oxford-aircraft}  &77.25\scriptsize{$\pm$0.09}&80.54\scriptsize{$\pm$0.08}&\textBF{80.67}\scriptsize{$\pm$0.18} \\ 
    Stanford Cars~\cite{stanford-cars} &82.13\scriptsize{$\pm$0.03}&90.20\scriptsize{$\pm$0.01}&\textBF{90.27}\scriptsize{$\pm$0.02}\\ 
        \bottomrule[1.5pt]
    \end{tabular}}
    \label{supp:fine}
\end{table}
To comprehensively assess the effectiveness of our proposed methods, we apply our proposed methods to various fine-grained datasets, including Oxford Flowers~\cite{oxford-flower}, Oxford-IIIT Pets~\cite{oxford-pets}, FGVC-Aircraft~\cite{oxford-aircraft}, and Stanford Cars~\cite{stanford-cars}.
Specifically, for all the fine-grained datasets, we employ the ResNet-50 model~\cite{resnet} pre-trained on ImageNet, followed by fine-tuning these models using our proposed methods.
To ensure fairness, experiments on the same dataset utilize the same experimental settings.
As shown in Table~\ref{supp:fine}, EntAugment brings notable accuracy improvements across all fine-grained datasets, and the incorporation of EntLoss further bolsters the performance gains.
Hence, the proposed methods can also be utilized to enhance the model performance on fine-grained datasets.
\end{document}